\pdfoutput=1
\documentclass[11pt]{article}

\usepackage[final]{acl}

\usepackage{times}
\usepackage{latexsym}

\usepackage[T1]{fontenc}
\usepackage[utf8]{inputenc}

\usepackage{microtype}
\usepackage{enumitem}

\usepackage{inconsolata}
\usepackage{tikz}
\usepackage[edges]{forest}
\definecolor{hiddendraw}{RGB}{205, 44, 36}
\definecolor{hidden-blue}{RGB}{99,178,238} % 194,232,247
\definecolor{hidden-orange}{RGB}{243,202,120}
\definecolor{hidden-yellow}{RGB}{253,225,28}
\definecolor{hidden-green}{RGB}{77,214,12}
\definecolor{hidden-grey}{RGB}{175,171,171}
\usepackage{amsmath}
\usepackage{amsfonts,amssymb, bbm}
\usepackage{booktabs}
\usepackage{multirow}
\usepackage{CJKutf8}
\usepackage{pifont} % 提供 \ding 命令
\usepackage{xcolor} % 提供 \color 命令

\usepackage{CJKutf8}

\newcommand{\cmark}{\color{green}\ding{51}}%
\newcommand{\xmark}{\color{red}\ding{55}}%

\title{A Survey of Ontology Expansion for Conversational Understanding}

\author{Jinggui Liang${}^{1}$, ~
        Yuxia Wu${}^{1}$, ~
        Yuan Fang${}^{1}$, ~
        Hao Fei${}^{2}$, ~
        Lizi Liao${}^{1}$\\
        ${}^{1}$Singapore Management University,
        ${}^{2}$National University of Singapore
        \\
        \texttt{jg.liang.2023@phdcs.smu.edu.sg} \quad \texttt{yieshah2017@gmail.com}  \\
        \texttt{yfang@smu.edu.sg} \quad \texttt{haofei37@nus.edu.sg} \quad \texttt{lzliao@smu.edu.sg}
        }

\begin{document}
\maketitle
\begin{abstract}

In the rapidly evolving field of conversational AI, \textbf{On}tology \textbf{Exp}ansion (\textbf{OnExp}) is crucial for enhancing the adaptability and robustness of conversational agents. Traditional models rely on static, predefined ontologies, limiting their ability to handle new and unforeseen user needs. This survey paper provides a comprehensive review of the state-of-the-art techniques in OnExp for conversational understanding. It categorizes the existing literature into three main areas: (1) \textit{New Intent Discovery}, (2) \textit{New Slot-Value Discovery}, and (3) \textit{Joint OnExp}. By examining the methodologies, benchmarks, and challenges associated with these areas, we highlight several emerging frontiers in OnExp to improve agent performance in real-world scenarios and discuss their corresponding challenges. This survey aspires to be a foundational reference for researchers and practitioners, promoting further exploration and innovation in this crucial domain.

\end{abstract}

\section{Introduction}
\label{ref:intro}

%Conversational understanding (CU) is a core component in developing conversational agents \citep{AliMe, Alexa}. The goal of the CU module is to accurately capture the underlying needs conveyed by users during their interactions with agents. As shown in Figure \ref{fig:loop}, the understanding capabilities of these agents are generally encapsulated within a conversational ontology, which defines a collection of all possible user intents, slots, and values for each slot \citep{ontology, budzianowski-etal-2018-multiwoz-1, ontology-2}. Effective CU models are tasked with not only identifying the overall purposes (\textit{intents detection}) \citep{intent_dec_1} expressed by users but also pinpointing relevant pieces of information (\textit{slots-filling}) \citep{intent_dec_2} that fulfill their intents.

\begin{figure}[!t]
\centering
\includegraphics[width=1.0\linewidth]{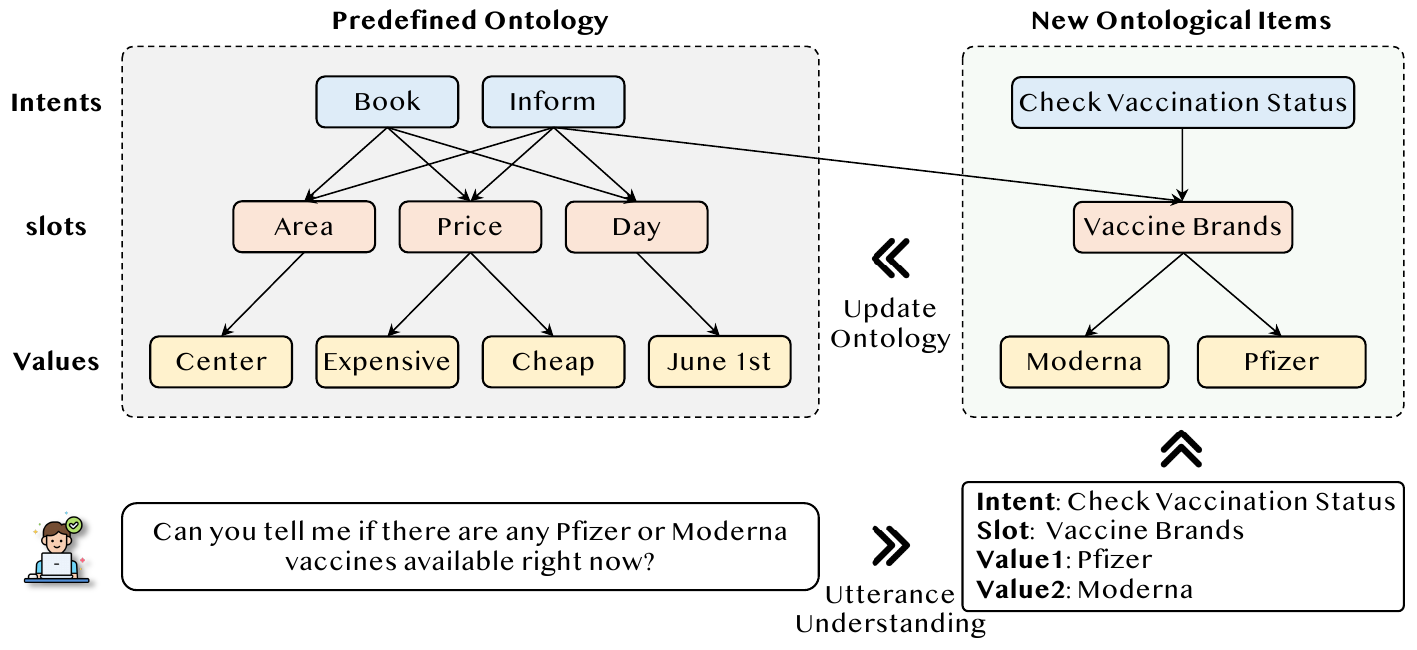} 
% \vspace{-0.7cm}
\caption{An example of ontology expansion enabling conversational agents to adapt to unseen events.
}
\label{fig:loop}
% \vspace{-0.5cm}
\end{figure}

Conversational understanding (CU) is a core component in the development of conversational agents \citep{AliMe, Alexa}. The objective of the CU module is to accurately capture and interpret user needs during interactions. As illustrated in Figure \ref{fig:loop}, these capabilities are generally encapsulated within a conversational ontology, which defines a collection of possible user intents, slots, and values for each slot \citep{ontology, budzianowski-etal-2018-multiwoz-1, ontology-2}. Effective CU models must not only identify the overall purposes (\textit{intent detection}) \citep{intent_dec_1} expressed by users but also pinpoint relevant pieces of information (\textit{slot filling}) \citep{intent_dec_2} that fulfill these intents.

%Typically, traditional CU research assumes that the conversational ontology is well-defined in advance, where all intents, slots, and values are given. With this predefined ontology, CU is approached as a close-word classification problem, wherein the task involves classifying user utterances into their respective ontological categories. However, as highlighted in Figure \ref{fig:loop}, conversational agents in real-world settings may encounter rapidly evolving user needs and a broad diversity of expressions, leading to the ongoing emergence of new ontological items within input utterances \citep{supnid_slm_13, supnid_slm_11}. This dynamic environment inevitably presents a significant challenge, as traditional CU models fall short when handling unseen situations not covered by the predefined ontology.
Traditionally, CU research assumes a well-defined, static ontology where all intents, slots, and most possible values are predetermined. Within this predefined framework, CU is often treated as a closed-world classification task for intents and sequence labeling task for slot values \citep{DBLP:journals/corr/abs-2207-13211}. However, in real-world settings, conversational agents encounter rapidly evolving user needs and diverse expressions, leading to the emergence of new ontological items \citep{supnid_slm_13, supnid_slm_11}. This dynamic environment presents a significant challenge, as traditional CU models fail easily in situations beyond the predefined ontology.

%Addressing this challenge, \textbf{On}tology \textbf{Exp}ansion (\textbf{OnExp}) has been proposed and extensively investigated to facilitate open-world ontology learning \citep{supnid_slm_5, supnid_slm_6, textoir, supnid_slm_8}. At its core, OnExp seeks to dynamically update and extend the conversational ontology by recognizing both pre-established and novel ontological items that emerge from user utterances. Proficient OnExp approaches can not only improve the downstream decision-making and policy implementation of conversational agents but also significantly boost user satisfaction and service efficiency.
To address this challenge, OnExp has been proposed to facilitate open-world ontology learning \citep{supnid_slm_5, supnid_slm_6, textoir, supnid_slm_8,wu2022semi}. It dynamically updates and extends the conversational ontology by recognizing both pre-established and novel ontological items from user utterances. Effective OnExp approaches can significantly enhance the downstream decision-making and policy implementation of conversational agents, improving user satisfaction and service efficiency \citep{dao-etal-2023-reinforced, DBLP:conf/sigir/Dao0LL24}.

%In recent years, significant research progress has been made in developing innovative and effective OnExp methodologies. However, due to the rapid advancements in the field, there is a notable lack of a comprehensive review summarizing such efforts and discussing the emerging trends. This absence impedes the further development of OnExp. Therefore, this paper aims to fill this gap by conducting the first comprehensive and well-structured survey of this research field. Specifically, we initially introduce the preliminaries of OnExp, detailing the relevant task formulations, commonly used data resources, and evaluation protocols. We then present a novel taxonomy to systematically organize the OnExp studies within the literature from a unified perspective. According to the types of ontological items involved in the model learning, we categorize OnExp methods into three categories: (1) \textit{New Intent Discovery} (NID), (2) \textit{New Slot-Value Discovery} (NSVD), and (3) \textit{Joint OnExp}. This taxonomy can provide comprehensive coverage of all types of OnExp, aiding researchers from various fields in thoroughly tracking recent breakthroughs. Finally, we offer a detailed outlook on several promising research directions for future OnExp studies and discuss their corresponding challenges, aiming to motivate further research.

\begin{table*}[t!]
\centering
\resizebox{1.0\linewidth}{!}{
\begin{tabular}{lcccccc}
\toprule
\multirow{2}{*}[-1pt]{\textbf{Dataset}} & \multirow{2}{*}[-1pt]{\textbf{Domain}} & \multirow{2}{*}[-1pt]{\textbf{\#Samples}} & \multicolumn{2}{c}{\textbf{\#Ontologies}} & \multicolumn{2}{c}{\textbf{Supported Tasks}} \\ \cline{4-7}
    &   &   &           &         &     &  \\[-11.5pt]
    &   &   & \#Intents  & \#Slots  & NID & NSVD \\
\midrule
BANKING77 \citep{banking_dataset}             & Bank         & 13,083  & 77  & -  & \cmark & \xmark \\
CLINC150 \citep{clinc_dataset}                & Multi-domain &  22,500 & 150 & -  & \cmark & \xmark \\
StackOverflow \citep{stackoverflow_dataset}   & Question     &  20,000 & 20  & -  & \cmark & \xmark \\
\midrule
CamRest \citep{camrest_dataset}               & Restaurant   & 2,744   & 2   & 4  & \xmark & \cmark \\
Cambridge SLU \citep{canbridge_slu_dataset}   & Restaurant   & 10,569  & 5   & 5  & \xmark & \cmark \\
WOZ-attr \citep{multiwoz/EricGPSAGKGKH20}            & Attraction   & 7,524   & 3   & 8  & \xmark & \cmark \\
WOZ-hotel \citep{multiwoz/EricGPSAGKGKH20}           & Hotel        & 14,435  & 3   & 9  & \xmark & \cmark \\
ATIS \citep{atis_dataset}                     & Flight       &  4,978  & 17  & 79 & \cmark & \cmark \\
SNIPS \citep{Snips_dataset}                   & Multi-domain & 13,784  & 7   & 72  & \cmark & \cmark \\
SGD \citep{SGD/RastogiZSGK20}                 & Multi-domain & 329,964  & 46   & 214  & \cmark & \cmark \\
\bottomrule
\end{tabular}
}
% \vspace{-0.1cm}
\caption{Summary of popular datasets for OnExp. \#Samples, \#Intents, and \#Slots represent the total number of utterances, intents, and slots, respectively.}
\label{tab:datasets}
% \vspace{-0.5cm}
\end{table*}

Recent years have witnessed substantial progress in developing innovative OnExp methodologies. However, the rapid advancements have left a gap in comprehensive reviews that summarize these efforts and discuss emerging trends. This paper aims to fill this gap by providing a thorough survey of OnExp research. We introduce the preliminaries of OnExp, detailing task formulations, data resources, and evaluation protocols. Our novel taxonomy categorizes OnExp studies into three types: (1) New Intent Discovery (NID), (2) New Slot-Value Discovery (NSVD), and (3) Joint OnExp, offering comprehensive coverage of the field. Finally, we discuss promising research directions and associated challenges, motivating further exploration.

In summary, our contributions are as follows:(1) We present the first comprehensive survey on ontology expansion; (2) We categorize OnExp research into three branches: NID, NSVD, and Joint OnExp, providing a unified understanding of the literature; (3) We discuss emerging frontiers and challenges in OnExp, highlighting future research directions. Additionally, we maintain a GitHub repository\footnote{\url{https://github.com/liangjinggui/Ontology-Expansion}} that organizes useful resources.

\section{Preliminaries}

% To better understand the state and advancements that have shaped the field of ontology expansion, 

%In this section,  we first describe the formal formulation for OnExp tasks (\S \ref{pre:task formailzation}), then review the existing data resources widely used in this field (\S \ref{pre:data resources}). Finally, we present the corresponding evaluation protocols for OnExp (\S \ref{pre:eval protocols}).

\subsection{Task Formulation}
\label{pre:task formailzation}

Ontology expansion in conversational understanding involves dynamically broadening the predefined ontology by recognizing both known and novel ontological items from user utterances. These items are structured as a collection of intents, slots, and corresponding slot values.

Formally, let $\mathcal{O}_k$ and $\mathcal{O}_u$ represent the sets of predefined and unknown ontological items, with $ \mathcal{O}_u \cap \mathcal{O}_k = \varnothing$. The OnExp tasks consider a dataset \(\mathcal{D}^{all}\) that is divided into two parts: a labeled dataset \(\mathcal{D}^{l}\) and an unlabeled dataset \(\mathcal{D}^{u}\). \(\mathcal{D}^{l} = \{(\boldsymbol{x}_i, o_i) | o_i \in \mathcal{O}_k\}_{i=1}^{\lvert \mathcal{D}^{l} \rvert}\) consists of utterances paired with labels that belong to \(\mathcal{O}_k\). Conversely, \(\mathcal{D}^{u} = \{\boldsymbol{x}_i | o_i \in \mathcal{O}_u \cup \mathcal{O}_k \}_{i=1}^{\lvert \mathcal{D}^{u} \rvert}\) includes utterances for which the labels are not available during the model learning, covering both $\mathcal{O}_k$ and $\mathcal{O}_u$.

Given an utterance $\boldsymbol{x}_i \in \mathcal{D}^{all}$, the overall objective of OnExp tasks is to optimize a mapping function \( f_{\theta}^{OnExp} \), parameterized by \( \theta \), to recognize its corresponding ontological items as follows:
\begin{equation}
    f_{\theta}^{OnExp}(\boldsymbol{x}_i) \rightarrow (o_i^{I}, o_i^{S}, o_i^{V}, r),
\end{equation}
where $(o_i^{I}, o_i^{S}, o_i^{V}) \in \mathcal{O}_k \cup \mathcal{O}_u$ denote the intent, slot, and value associated with $\boldsymbol{x}_i$. The term $r$ refers to the relations among various ontological items, such as the intent \textit{Check Vaccination Status} being associated with the slot \textit{Vaccine Brands}, but not with the slot \textit{Area}. As the focus of OnExp is on identifying and expanding fundamental concepts emerging from dynamic conversations, the relations among these items are typically overlooked in the existing literature.

As discussed in Section \ref{ref:intro}, OnExp encompasses various tasks. In the NID setting, the mapping function \( f_{\theta}^{OnExp} \) predicts only \( o^I \), discarding \( (o^S, o^V) \). In the NSVD setting, the focus shifts to uncovering \( (o^S, o^V) \), omitting intents \( o^I \). In Joint OnExp, \( (o^{I}, o^{S}, o^{V}) \) are all retained, with the aim of leveraging shared knowledge across these tasks for more effective ontology learning.

\subsection{Data Resources}
\label{pre:data resources}
High-quality annotated datasets are essential for developing OnExp methods. We summarize the commonly used data resources, with an overview of each dataset's domain, scale, annotated ontological items, and supported tasks in Table \ref{tab:datasets}. 

For \textbf{NID}, the most widely used datasets are BANKING77 \citep{banking_dataset}, CLINC150 \citep{clinc_dataset}, and StackOverflow \citep{stackoverflow_dataset}. For \textbf{NSVD}, prominent datasets include CamRest \citep{camrest_dataset}, Cambridge SLU \citep{canbridge_slu_dataset}, WOZ-attr \citep{multiwoz/EricGPSAGKGKH20}, WOZ-hotel \citep{multiwoz/EricGPSAGKGKH20}, ATIS \citep{atis_dataset}, SNIPS \citep{Snips_dataset}, and SGD \citep{SGD/RastogiZSGK20}. Further details on these datasets are provided in Appendix \ref{app:data_resources}.

%High-quality annotated datasets are essential for developing OnExp methods. This section summarizes the commonly used data resources for OnExp. An overview of each dataset, detailing its domain, scale, annotated ontological items, and supported tasks, is presented in Table \ref{tab:datasets}. For clarity, we organize these datasets into the following categories:

%\vspace{0.2em}
%\noindent
%\textbf{NID Datasets.} The most widely used datasets for NID are: (1) \textbf{BANKING77} \citep{banking_dataset}; (2) \textbf{CLINC150} \citep{clinc_dataset}; and (3) \textbf{StackOverflow} \citep{stackoverflow_dataset}.
%\textbf{NSVD Datasets.} We introduce six prominent datasets spanning various domains for the NSVD task: (1) \textbf{CamRest} \citep{camrest_dataset}; (2) \textbf{Cambridge SLU} \citep{canbridge_slu_dataset}; (3) \textbf{WOZ-attr} \citep{multimoz_dataset2}; (4) \textbf{WOZ-hotel} \citep{multimoz_dataset1}; (5) \textbf{ATIS} \citep{atis_dataset}; (6) \textbf{SNIPS} \citep{Snips_dataset}.

%The details of these datasets are presented in Appendix \ref{app:data_resources}. 

\subsection{Evaluation Protocols}
\label{pre:eval protocols}
%Here, we discuss the evaluation metrics applicable to the included datasets when evaluating the performance of OnExp methods. 
% In particular, we present these metrics accordingly, based on the previously discussed task settings.
%\vspace{0.2em}

\paragraph{NID Metrics.} The NID evaluation metrics include: (1) Accuracy (\textbf{ACC}), based on the Hungarian algorithm; (2) Adjusted Rand Index (\textbf{ARI}); and (3) Normalized Mutual Information (\textbf{NMI}).

\paragraph{NSVD Metrics.} The performance of NSVD systems is evaluated using the following key metrics: (1) \textbf{Precision}, (2) \textbf{Recall}, and (3) \textbf{F1-score}. The F1-score, which is calculated based on slot value spans, is also referred to as \textbf{Span-F1}.

\paragraph{Other Metrics.} Notably, these evaluation metrics are not confined to the corresponding settings described previously. Additionally, the OnExp models can also be evaluated by \textbf{Known Acc}, \textbf{Novel Acc}, and \textbf{H-score} \citep{supnid_slm_11}. Thorough discussions and specific definitions of the above evaluation metrics are detailed in Appendix \ref{app:eval_protocols}.

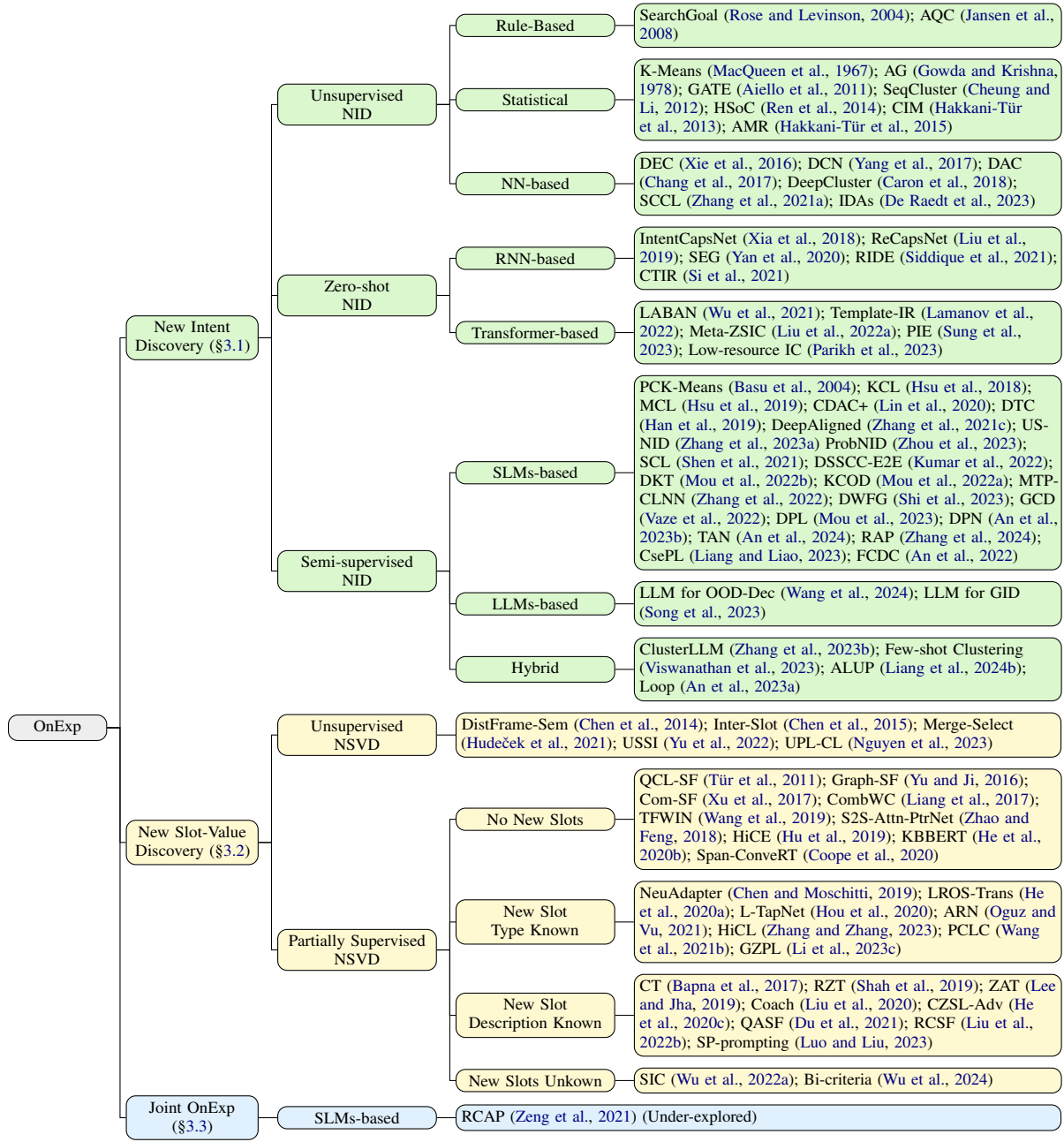
\begin{figure*}[!t]
	\scriptsize
	\begin{forest}
		for tree={
			forked edges,
			grow'=0,
			draw,
			rounded corners,
			node options={align=center},
			calign=edge midpoint,
		},
	    [OnExp, text width=1.3cm, for tree={fill=hidden-grey!20}
			[New Intent Discovery (\S \ref{tax:NID}), text width=1.8cm, for tree={fill=hidden-green!20}
                    [Unsupervised\\NID, text width=2.2cm, for tree={fill=hidden-green!20}
    				[Rule-Based, text width=2.2cm, for tree={fill=hidden-green!20}
                            [SearchGoal \citep{unsupnid_rule_1};
                             AQC \citep{unsupnid_rule_2}, 
                              text width=6.2cm, node options={align=left}
                            ]
                        ]
                        [Statistical, text width=2.2cm, for tree={fill=hidden-green!20}
                            [K-Means \citep{unsupnid_stat_6};
                             AG \citep{unsupnid_stat_7};
                             GATE \citep{unsupnid_stat_1};
                             SeqCluster \citep{unsupnid_stat_2};
                             HSoC \citep{unsupnid_stat_3};
                             CIM \citep{unsupnid_stat_4};
                             AMR \citep{unsupnid_stat_5}, 
                             text width=6.2cm, node options={align=left}
                            ]
                        ]
                        [NN-based, 
                         text width=2.2cm, for tree={fill=hidden-green!20}
                            [DEC \citep{unsupnid_nn_1}; 
                             DCN \citep{unsupnid_nn_2}; 
                             DAC \citep{unsupnid_nn_3}; 
                             DeepCluster \citep{unsupnid_nn_4}; 
                             SCCL \citep{unsupnid_nn_5}; 
                             IDAs \citep{unsupnid_nn_6}, 
                             text width=6.2cm, node options={align=left}
                            ]
                        ] 
				]
                [Zero-shot\\NID, text width=2.2cm, for tree={fill=hidden-green!20}
                    [RNN-based, text width=2.2cm, for tree={fill=hidden-green!20}
                        [
                         IntentCapsNet \citep{emnlp/XiaZYCY18};
                         ReCapsNet \citep{emnlp/LiuZFFLWL19};
                         SEG \citep{acl/YanFLLZWL20};
                         RIDE \citep{DBLP:conf/sigir/SiddiqueJXH21};
                         CTIR \citep{ijcai/SiL00LW21},
                         text width=6.2cm, node options={align=left}
                        ]
                    ]
                    [Transformer-based, text width=2.2cm, for tree={fill=hidden-green!20}
                        [
                         LABAN \citep{emnlp/WuSJ21};
                         Template-IR \citep{lamanov-2022-template};
                         Meta-ZSIC \citep{sigir/LiuZZZS0Z22};
                         PIE \citep{emnlp/SungGM0SRZC23};
                         Low-resource IC \citep{acl/ParikhTTV23},
                         text width=6.2cm, node options={align=left}
                        ]
                    ]
                ]  
                 [Semi-supervised\\NID, text width=2.2cm, for tree={fill=hidden-green!20}
                    [SLMs-based, text width=2.2cm, for tree={fill=hidden-green!20}
                        [PCK-Means \citep{supnid_slm_1}; 
                         KCL \citep{supnid_slm_2}; 
                         MCL \citep{supnid_slm_3}; 
                         CDAC+ \citep{supnid_slm_5}; 
                         DTC \citep{supnid_slm_4};  
                         DeepAligned \citep{supnid_slm_6}; 
                         USNID \citep{supnid_slm_6.5}
                         ProbNID \citep{supnid_slm_9};
                         SCL \citep{supnid_slm_7};
                         DSSCC-E2E \citep{supnid_slm_7.5};
                         DKT \citep{supnid_slm_7.8};
                         KCOD \citep{supnid_slm_7.7};
                         MTPCLNN \citep{supnid_slm_8};
                         DWFG \citep{supnid_slm_14}; 
                         GCD \citep{supnid_slm_7.6};
                         DPL \citep{supnid_slm_9.5};
                         DPN \citep{supnid_slm_10};
                         TAN \citep{supnid_slm_11};
                         RAP \citep{supnid_slm_12};
                         CsePL \citep{supnid_slm_13};
                         FCDC \citep{supnid_slm_15},
                         text width=6.2cm, node options={align=left}]
                    ]
                    [LLMs-based, text width=2.2cm, for tree={fill=hidden-green!20}
                        [LLM for OOD-Dec \citep{supnid_llm_1};
                         LLM for GID \citep{supnid_llm_2},
                         text width=6.2cm, node options={align=left}
                        ]
                    ]
                    [Hybrid, text width=2.2cm, for tree={fill=hidden-green!20}
                        [ClusterLLM \citep{supnid_hybrid_1}; 
                         Few-shot Clustering \citep{supnid_hybrid_2}; 
                         ALUP \citep{supnid_hybrid_4}; 
                         Loop \citep{supnid_hybrid_3},
                         text width=6.2cm, node options={align=left}
                        ]
                    ]
                ]  
			]
			[New Slot-Value Discovery (\S \ref{tax:NSVD}), text width=1.8cm, for tree={fill=hidden-yellow!20}
                        [Unsupervised\\NSVD, text width=2.2cm, for tree={fill=hidden-yellow!20}
    					[DistFrame-Sem \citep{chen2014leveraging};
                             Inter-Slot \citep{chen2015jointly};
                             Merge-Select \citep{hudevcek2021discovering};
                             USSI \citep{yu2022unsupervised}; 
                             UPL-CL \citep{nguyen-etal-2023-slot},
    					text width=8.85cm, node options={align=left}
    					]
                         ]
				   % [\ljg{Hold Unsupervised\\NSVD}, text width=2.2cm, for tree={fill=hidden-yellow!20}
       %                  [NVD-based, text width=2.2cm, for tree={fill=hidden-yellow!20}
       %                   [QCL-SF \citep{unsup_nvd_1};
       %                    Graph-SF \citep{unsup_nvd_2};
       %                    Com-SF \citep{community-based-SF}
       %                    TFWIN \citep{TFWIN}, 
       %                   text width=6.2cm, node options={align=left}
       %                   ]
       %                   ]
       %                  [NSD-based, text width=2.2cm, for tree={fill=hidden-yellow!20}
       %                   [DistFrame-Sem \citep{chen2014leveraging};
       %                    Inter-Slot \citep{chen2015jointly};
       %                    Merge-Select \citep{hudevcek2021discovering};USSI \citep{yu2022unsupervised}; UPL-CL \citep{nguyen-etal-2023-slot},  
       %                   text width=6.2cm, node options={align=left}
       %                   ]
       %                   ]
       %                ]                                     
                    [Partially Supervised\\NSVD , text width=2.2cm, for tree={fill=hidden-yellow!20}	
                    [No New Slots, text width=2.2cm, for tree={fill=hidden-yellow!20}
                    [QCL-SF \citep{unsup_nvd_1};
                     Graph-SF \citep{unsup_nvd_2};
                     Com-SF \citep{community-based-SF};
                     CombWC \citep{nvd_based_1};
                     TFWIN \citep{TFWIN};
                     S2S-Attn-PtrNet \citep{nvd_based_2};
                     HiCE \citep{nvd_based_6};
                     KBBERT \citep{nvd_based_4};
                     Span-ConveRT \citep{nvd_based_5}, 
                     text width=6.2cm, node options={align=left}
                    ]
                   ]  
                     [New Slot Type Known, text width=2.2cm, for tree={fill=hidden-yellow!20}
                         [
                          NeuAdapter \cite{nvd_based_8};
                          LROS-Trans \citep{nvd_based_9};
                          L-TapNet \citep{L-TapNet/acl/HouCLZLLL20};
                          ARN \citep{ARN/eacl/OguzV21};
                          HiCL \citep{HiCL};
                          PCLC \citep{PCLC};
                          GZPL \citep{GZPL},  
                         text width=6.2cm, node options={align=left}
                         ]
                         ]
                     [New Slot \\Description Known, text width=2.2cm, for tree={fill=hidden-yellow!20}
                      [
                       CT \citep{semi_nvd_based_1};
                       RZT \citep{nvd_based_7};
                       ZAT \citep{zat/aaai/LeeJ19};
                       Coach \citep{Coach};
                       CZSL-Adv \citep{CZSL-Adv};
                       QASF \citep{QASF}; 
                       RCSF \citep{RCSF};
                       SP-prompting \citep{SP-prompting}
                       , 
                       text width=6.2cm, node options={align=left}
                      ]
                     ]
                    [New Slots Unkown, text width=2.2cm, for tree={fill=hidden-yellow!20}
                    [SIC \citep{wu2022semi};
                     Bi-criteria \citep{wu2024active}, 
                     text width=6.2cm, node options={align=left}
                    ]
                   ]  
                  ]
			]
                [Joint OnExp \\(\S \ref{tax:Joint ProOE}), text width=1.8cm, for tree={fill=hidden-blue!20}
                [SLMs-based, text width=2.2cm, for tree={fill=hidden-blue!20}
					[RCAP \citep{joint_prooe_1} (Under-explored),
					text width=8.85cm, node options={align=left}
					]
                ]		
			]   
	]
	\end{forest}
        % \vspace{-0.2cm}

	\caption{The taxonomy for Ontology Expansion.
 }
    \label{fig:taxonomy}
% \vspace{-0.3cm}
\end{figure*}

% For all the evaluation metrics mentioned above, specific definitions are detailed in Appendix \ref{app:eval_protocols}. 
%\yx{what are Known Accuracy, Novel Accuracy, and H-score}

\section{Taxonomy of OnExp Research}

This section presents the new taxonomy for OnExp as shown in Figure \ref{fig:taxonomy}, comprising \textit{New Intent Discovery} (\S \ref{tax:NID}), \textit{New Slot-Value Discovery} (\S \ref{tax:NSVD}), and \textit{Joint OnExp} (\S \ref{tax:Joint ProOE}).

\subsection{New Intent Discovery}
\label{tax:NID}
We first explore the NID task in this section, which aims to simultaneously identify known and newly emerged user intents. Notably, NID operates at the utterance level, excelling in isolating distinct user intents but struggling with overlapping or ambiguous ones. To achieve effective NID, a variety of methodologies have been devised, as illustrated in Figure \ref{fig:taxonomy}. We classify these NID studies into three categories based on the use of available labeled data: Unsupervised NID, Zero-shot NID, and Semi-supervised NID.

\subsubsection{Unsupervised NID}
Unsupervised NID aims to discover user intents without any labeled data, facing significant challenges in deriving effective intent patterns to group similar utterances. This section categorizes existing unsupervised NID efforts into three types based on their model designs: Rule-based, Statistical, and Neural Network-based (NN-based) Methods.

\paragraph{Rule-based Methods.} Early efforts, such as those by \citet{unsupnid_rule_1}, collaborated with domain experts to develop a conceptual schema for user goals, adapting to new goal categories. \citet{unsupnid_rule_2} used a decision tree for intent analysis. However, maintaining these rule-based models proved challenging as the complexity of rules intensified across different domains.
\label{unsup_nid:rule}

\paragraph{Statistical Methods.} Given the limitations inherent in rule-based systems, statistical methods emerged as a more robust and effective alternative. Typical clustering algorithms like K-Means \citep{unsupnid_stat_6} and Agglomerative Clustering \citep{unsupnid_stat_7} laid the groundwork. Later, \citet{unsupnid_stat_1} aggregated fine-grained intent-related missions to learn new search intents, while \citet{unsupnid_stat_2} used external knowledge bases for sequence clustering. Methods like \citet{unsupnid_stat_3} utilized heterogeneous graphs for cross-source intent learning, and \citet{unsupnid_stat_4} introduced Bayesian models leveraging clicked URLs as implicit supervision in clustering new intents, while \citet{unsupnid_stat_5} explored the lexical semantic structure of user utterances with semantic parsers. Despite their robustness, these methods often struggled with high-dimensional data and complex semantics.
\label{unsup_nid:stat}

\paragraph{NN-based Methods.} To address the limitations of statistical methods, deep neural models have been explored for more effective new intent learning, thanks to their superior learning capabilities and flexible parameters. \citet{unsupnid_nn_1} proposed Deep Embedded Clustering (DEC), which iteratively refines intent clusters using an auxiliary target distribution. \citet{unsupnid_nn_2} developed a Deep Clustering Network (DCN) that combines nonlinear dimensionality reduction with K-Means clustering to optimize utterance representations. Deep Adaptive Clustering (DAC) \citep{unsupnid_nn_3} reimagined intent discovery as a pairwise classification problem, employing a binary-constrained model to learn relationships between utterance pairs. DeepCluster \citep{unsupnid_nn_4} alternated between clustering utterances and refining their representations via cluster assignments. Further advancements include Supporting Clustering with Contrastive Learning (SCCL) \citep{unsupnid_nn_5}, which utilized emerging contrastive learning techniques to enhance intent clustering. In the era of Large Language Models (LLMs), \cite{unsupnid_nn_6,liang-etal-2024-synergizing} further leveraged LLMs to enhance intent clustering.
\label{unsup_nid:nn}

\subsubsection{Zero-shot NID.}

Zero-shot NID aims to discover new user intents using only labeled training data from known intents. The main challenge lies in effectively transferring the prior knowledge of known intents to facilitate the recognition of new intents. This setting is divided into RNN-based and Transformer-based methods based on their backbone architecture.

\paragraph{RNN-based Methods.} RNNs were the dominant model for encoding sentences in the early days. Hence, \citet{emnlp/XiaZYCY18} proposed an RNN-based capsule network with routing-by-agreement to adapt the model to new intents. To address the polysemy problem, \citet{emnlp/LiuZFFLWL19} introduced a dimensional attention mechanism and learned generalizable transformation matrices for new intents. Beyond merely extracting features from utterances, \citet{DBLP:conf/sigir/SiddiqueJXH21} incorporated commonsense knowledge to learn robust relationship meta-features. Despite these advancements, \citet{ijcai/SiL00LW21} identified a critical issue: new intent representations cannot be learned during training. Hence, they proposed the Class-Transductive Intent Representations framework, which progressively optimizes new intent features using intent names.

\paragraph{Transformer-based Methods.} In practice, the sequential nature of RNNs incurs high computational costs and struggles with long-range dependencies. To address these issues, Transformers have emerged as an effective solution for zero-shot NID. \citet{emnlp/WuSJ21} developed a label-aware BERT attention network that constructs an intent label semantic space to map utterances to intent labels. Following this, \citet{lamanov-2022-template} modeled this task as a sentence pair modeling problem, utilizing pre-trained language models to fuse intent labels and utterances for binary classification. \citet{sigir/LiuZZZS0Z22} introduced a mixture attention mechanism and collaborated it with a novel meta-learning paradigm to enhance new intent identification. To better adapt pre-trained encoders to intent discovery, \citet{emnlp/SungGM0SRZC23} proposed generating pseudo-intent names from utterances and applied intent-aware contrastive learning to develop the Pre-trained Intent-aware Encoder (PIE). Recently, \citet{acl/ParikhTTV23} explored zero-shot NID using Large Language Models (LLMs), investigating the various strategies such as in-context prompting to aid in identifying novel intents.

% LABAN \citep{emnlp/WuSJ21};
% Template-IR \citep{lamanov-2022-template};
% Meta-ZSIC \citep{sigir/LiuZZZS0Z22};
% PIE \citep{emnlp/SungGM0SRZC23};
% Low-resource IC \citep{acl/ParikhTTV23},

\subsubsection{Semi-supervised NID.}

Semi-supervised NID combines limited labeled data with extensive unlabeled data to discern new intents. This approach faces challenges in deriving supervision signals for unlabeled utterances and avoiding overfitting to known intents. Unlike Zero-shot NID, which is provided with new intent names or classes, semi-supervised NID does not know the new intents or their quantity. This section categorizes methods into Small Language Models (SLMs)-based, LLMs-based, and Hybrid methods.

\paragraph{SLMs-based Methods.} SLMs like BERT, pre-trained on large-scale corpora, exhibit strong text understanding abilities and have been effectively fine-tuned for various tasks \citep{bert/DevlinCLT19, bart/LewisLGGMLSZ20}. Utilizing SLMs as feature extractors, \citet{supnid_slm_1} introduced Pairwise Constrained K-Means (PCK-Means) with active constraint selection for new intent clustering. Building on this, \citet{supnid_slm_2} used SLMs for static constraints with KL divergence-based Contrastive Loss (KCL), while \citet{supnid_slm_3} proposed Meta Classification Likelihood (MCL) for dynamic pairwise similarity updates. \citet{supnid_slm_5} presented Constrained Deep Adaptive Clustering (CDAC+) for iterative model refinement.
\label{semi_sup_nid:slms}

Despite these advances, pairwise supervision signals often fall short in fully utilizing labeled data. To address this, \citet{supnid_slm_4} proposed Deep Transfer Clustering (DTC), improving clustering quality through consistency regulation and intent cluster number estimation. \citet{supnid_slm_6} developed DeepAligned to resolve label inconsistencies, later improved by USNID for faster convergence \citep{supnid_slm_6.5}. \citet{supnid_slm_9} alleviated prior knowledge forgetting with ProbNID, a probabilistic framework optimizing intent assignments via Expectation Maximization. \citet{supnid_slm_8} utilized multi-task pre-training and K-nearest neighbor contrastive learning for compact clusters (MTPCLNN). Additionally, \citet{supnid_slm_14} proposed the Diffusion Weighted Graph Framework (DWGF), capturing both semantic and structural relationships within utterances for more reliable supervisory signals.
Beyond learning contrastive relations, \citet{supnid_slm_10} formulated a bipartite matching problem, proposing the Decoupled Prototypical Network (DPN) to separate known from new intents, facilitating explicit knowledge transfer. \citet{supnid_slm_12} introduced Robust and Adaptive Prototypical learning (RAP) to enhance intra-cluster compactness and inter-cluster dispersion. Recently, \citet{supnid_slm_13} leveraged prompt learning with two-level contrastive learning and soft prompting for new intent discovery.

While successful, SLM-based methods require extensive fine-tuning on large datasets, which is time-consuming. Moreover, SLMs struggle to fully capture the nuanced semantics of diverse and dynamic human languages in conversational contexts.

\paragraph{LLMs-based Methods.} Recently, LLMs \citep{Achiam2023GPT4TR, touvron2023llama} have shown impressive efficacy across a broad range of NLP tasks, such as summarization \citep{Liu2023OnLT} and query rewriting \citep{Anand2023QueryUI, DBLP:conf/naacl/GuoLZWLC24}. Given the above SLMs' limitations, there is a growing trend toward using LLMs for intent discovery in few/zero-shot settings. \citet{supnid_llm_1} evaluated LLMs' ability to detect unknown intents, using ChatGPT to classify intents beyond the predefined set. Moreover, \citet{supnid_llm_2} broadened the use of LLMs in intent discovery, directing ChatGPT to group utterances and identify known and novel intents.
\label{semi_sup_nid:llms}

\paragraph{Hybrid Methods.} Although LLMs-based methods excel in zero-shot settings, they typically underperform compared to fully fine-tuned models. To address this, Hybrid methods that combine the strengths of SLMs and LLMs have been developed to enhance intent discovery. In this effort, \citet{supnid_hybrid_1} proposed ClusterLLM, which uses triplet feedback from LLMs to refine SLMs-learned representations and applies pairwise hierarchical clustering to improve cluster granularity. Further, \citet{supnid_hybrid_2} investigated three strategies—keyphrase expansion, pairwise constraints, and cluster correction—to leverage LLMs for better intent clustering. To effectively utilize LLMs and reduce costs, \citet{supnid_hybrid_4} integrated LLMs into active learning, using uncertainty propagation to selectively label utterances and extending this feedback without spreading inaccuracies. Similarly, \citet{supnid_hybrid_3} introduced local inconsistent sampling with scalable queries to correct inaccurately allocated utterances using LLMs.
\label{semi_sup_nid:hybrid}

\subsection{New Slot-Value Discovery}
\label{tax:NSVD}

The NSVD task seeks to identify new slots and the corresponding values that emerge from dynamic conversations. Unlike the previous NID task that focuses on utterance-level recognition, NSVD specifically narrows its scope within individual utterances, excelling in detailed information extraction but limited by the quality and specificity of input data. Innovations in this task can be classified into unsupervised NSVD and partially supervised NSVD.

\subsubsection{Unsupervised NSVD}
Unsupervised NSVD discovers new slots and values without any labeled data, facing challenges such as dialogue noise and requiring high human intervention for ranking or selection processes. Early works like \citet{Unsupervised_induction} combined a frame-semantic parser with a spectral clustering-based slot ranking model to induce semantic slots. \citep{chen2014leveraging} further refined this method by integrating semantic frame parsing with word embeddings. Moreover, \citet{chen2015jointly} enhanced slot discovery by constructing lexical knowledge graphs and employing random walks to delineate slots. Despite the benefits of linguistic tools for discovering new slots, such methods struggled with dialogue noise and the ranking processes require significant human intervention. Addressing these challenges, \citet{hudevcek2021discovering} revised the ranking method to iteratively refine the obtained slots through slot taggers. To reduce reliance on generic parsers, \citet{yu2022unsupervised} further proposed a unified slot schema induction method that incorporates data-driven candidate value extraction and coarse-to-fine slot clustering. Recently, \citet{nguyen-etal-2023-slot} utilized pre-trained language model probing combined with contrastive learning refinement to induce value segments for slot induction.

\subsubsection{Partially Supervised NSVD}
Partially supervised NSVD leverages some form of labeled data and is divided into four types based on the supervision nature: No New Slots, New Slot Type Known, New Slot Description Known, and New Slot Unknown.

\paragraph{No New Slots.} This setting operates with all slot types predefined and certain known values for each slot labeled. It primarily explores leveraging existing slots to identify new values within these predefined slots, facing challenges in efficiently mining new value entities and leveraging external knowledge. This is common in scenarios where new restaurant names or new vaccine brand names emerge. Specifically, \citet{unsup_nvd_1} mined new slot entities from user queries in query click logs with target URLs, while \citet{unsup_nvd_2} used dependency trees to identify slot-specific triggers. \citet{community-based-SF} introduced a slot filler refinement method that constructs entity communities to filter out incorrect new fillers. \citet{nvd_based_1} combined word/character-level embeddings via highway networks to detect new values. Further, \citet{TFWIN} explored the temporal slot-filling problem and proposed a pattern-based framework that assesses pattern reliability and detects conflicts to find temporal values. To tackle the unknown value issue more effectively, \citet{nvd_based_6} formulated a K-shot regression problem, using a hierarchical context encoder and meta-learning to better infer new value embeddings. To explore the potential of external knowledge in aiding the discovery of new values, \citet{nvd_based_4} employed background knowledge bases with a knowledge integration method to facilitate tagging slot values.

\begin{table*}[t]
\centering
%\footnotesize
\renewcommand*{\arraystretch}{1.0}
\resizebox{1.0\linewidth}{!}{
\begin{tabular}{c|lllllllll}
\toprule
\multirow{3}{*}[3pt]{Methods} & \multicolumn{3}{c}{\textbf{BANKING77}} & \multicolumn{3}{c}{\textbf{CLINC150}} & \multicolumn{3}{c}{\textbf{StackOverflow}} \\ \cline{2-10}
&         &         &         &          &         &         &           &           &           \\[-6pt]
& \textbf{ACC} & ARI & NMI & \textbf{ACC}& ARI & NMI & \textbf{ACC} & ARI & NMI       \\
\midrule
\multicolumn{10}{c}{\textit{SLMs-based Methods}} \\ % 添加空白行
\midrule
PCK-Means \citep{supnid_slm_1} & 32.66 & 16.24 & 48.22 & 54.61 & 35.40 & 68.70 & 24.16 & 5.35 & 17.26 \\
BERT-KCL \citep{supnid_slm_2}  & 60.15 & 46.72 & 75.21 & 68.86 & 58.79 & 86.82 & 13.94 & 7.81 & 8.84 \\
BERT-MCL  \citep{supnid_slm_3} & 61.14 & 47.43 & 75.68 & 69.66 & 59.92 & 87.72 & 72.07 & 57.43 & 66.81 \\
CDAC+  \citep{supnid_slm_5}    & 53.83 & 40.97 & 72.25 & 69.89 & 54.33 & 86.65 & 73.48 & 52.59 & 69.84 \\
BERT-DTC \citep{supnid_slm_4}  & 56.51 & 44.70 & 76.55 & 74.15 & 65.02 & 90.54 & 71.47 & 53.66 & 63.17 \\
% GCD  \citep{supnid_slm_7.6}    & 58.95 & 46.87 & 77.86 & 77.50 & 67.44 & 91.13 & 67.71 & 47.70 & 64.74 \\
DeepAligned \citep{supnid_slm_6}& 64.90 & 53.64 & 79.56 & 86.49 & 79.75 & 93.89 & - & - & - \\
MTPCLNN \citep{supnid_slm_8}   & 73.98 & 63.10 & 84.22 & 88.25 & 84.77 & 94.88 & 83.18 & 69.50 & 77.03 \\
ProbNID  \citep{supnid_slm_9}  & 74.03 & 62.92 & 84.02 & 88.99 & 83.00 & 95.01 & 80.50 & 65.70 & 77.32 \\
DPN   \citep{supnid_slm_10}    & 74.45 & 63.26 & 84.31 & 89.22 & 84.30 & 95.14 & 84.59 & 70.27 & 79.89 \\
RAP   \citep{supnid_slm_12}    & 76.27 & 65.79 & 85.16 & 91.24 & 86.28 & 95.93 & 86.60 & 71.73 & 82.36 \\
USNID \citep{supnid_slm_6.5}   & 78.36 & 69.54 & 87.41 & 90.36 & 86.77 & 96.42 & 85.66 & 74.90 & 80.13 \\
DFWG   \citep{supnid_slm_14}   & 79.38 & 68.16 & 86.41 & 94.49 & 90.05 & 96.89 & 87.60 & 75.30 & 81.73 \\
CsePL  \citep{supnid_slm_13}   & 81.93 & 71.36 & 87.70 & 93.46 & 88.88 & 96.58 & 87.80 & 75.99 & 82.81 \\
\midrule
\multicolumn{10}{c}{\textit{LLMs-based Methods}} \\ % 添加空白行
\midrule
LLM for GID \citep{supnid_llm_2}  & 64.22 & - & - & 84.33 & - & - & - & - & - \\
\midrule
\multicolumn{10}{c}{\textit{Hybrid Methods}} \\ % 添加空白行
\midrule
Few-shot Clustering \citep{supnid_hybrid_2} & 65.30 & - & 82.40 & 79.40 & - & 92.60 & - & - & - \\
ClusterLLM   \citep{supnid_hybrid_1}  & 71.20 & - & 85.15 & 83.80 & - & 94.00 & - & - & - \\
ALUP   \citep{supnid_hybrid_4}  & 82.85 & 73.10 & 88.35 & 94.93 & 89.22 & 97.43 & 87.70 & 76.03 & 83.14 \\
\bottomrule
\end{tabular}
} 
% \vspace{-0.1cm}
\caption{The main semi-supervised NID results on three benchmarks.}
% \lz{NID的 zero-shot的放appendix吧，应该再加一个NSVD的表，数据集跟NID不一样也没关系，如果可以, 附件里也加上NVSD的zero shot结果.或者是正文里放NID zero-shot(unsupervise)和semi-supervise的结果，NVSD的都放附件。unsupervised和semisupervised里面都有几类，表格里的方法最好分一下，KIR没必要放表里，caption里denote下就可以了}

\label{tab:nid_main}
% \vspace{-0.5cm}
\end{table*}

\paragraph{New Slot Type Known.} Unlike merely identifying new values for predefined slots, practical applications may require models to extract values for well-defined slots not seen during training. The main challenge is adapting models to new slots. To address this, \citet{nvd_based_8} explored transfer learning for labeling new values and developed a neural adapter to adapt previously trained models to these new slots. Further, \citet{nvd_based_9} improved transfer learning efficiency by learning the label-relational output structure to capture slot label correlations, while \citet{PCLC} introduced prototypical contrastive learning with label confusion to refine slot prototypes dynamically. Beyond using coarse slot label information, \cite{HiCL} introduced Hierarchical Contrastive Learning (HiCL), where coarse and fine-grained slot labels serve as supervised signals to assist in extracting cross-domain slot fillers. Recently, \citet{GZPL} explored advanced prompting techniques for identifying new values, using slot types and inverse prompting to enhance model performance.

\paragraph{New Slot Description Known.} In contrast to accessing well-defined new slot types, this setting deals with extracting new values using only coarse-grained descriptions of new slots. Concretely, %\citet{semi_nvd_based_1} proposed Concept Tagger (CT) which utilizes slot descriptions to boost the cross-domain slot-filling. \citet{nvd_based_7} incorporated existing slot descriptions along with slot values to improve slot representations when extracting new values. 
\citet{semi_nvd_based_1} proposed Concept Tagger (CT) for cross-domain slot-filling with slot descriptions, while \citet{nvd_based_7} used slot descriptions to improve slot representations. In addition, \citet{Coach} proposed a coarse-to-fine (Coach) method that initially learns value patterns coarsely, then fills them into fine slot types based on the similarity with the representation of each slot type description. Inspired by this, \citet{CZSL-Adv} enhanced Coach with contrastive loss and adversarial attacks to improve robustness. Contrary to previous methods, \citet{QASF} and \citet{RCSF} tackle the slot-filling problem as a reading comprehension task, extracting new values by answering questions derived from slot descriptions. Recently, \citet{SP-prompting} combined learnable prompt tokens and discrete tokens of slot descriptions to identify new values.

\begin{table*}[t]
\centering
\footnotesize
\renewcommand*{\arraystretch}{1.0}
\resizebox{1.0\linewidth}{!}{
\begin{tabular}{c|ccccc}
\toprule
Methods & \textbf{CamRest} & \textbf{Cambridge SLU} & \textbf{WOZ-hotel} & \textbf{WOZ-attr} & \textbf{ATIS} \\ 
\midrule
% BERT-KCL \citep{supnid_slm_2}  & 18.9 & 13.1 & 17.8 & 56.0 & 49.2 \\
% BERT-MCL \citep{supnid_slm_3} & 18.8 & 12.9 & 17.9 & 53.2 & 50.4 \\
CDAC+ \citep{supnid_slm_5}  & 20.4 & 17.8 & 17.4 & 55.2 & 58.2 \\
BERT-DTC \citep{supnid_slm_4}  & 13.1 & 13.8 & 17.0 & 54.5 & 54.3 \\
DeepAligned \citep{supnid_slm_6} & 66.3 & 63.3 & 37.8 & 64.4 & 62.9 \\
SIC \citep{wu2022semi}  & 70.6 & 77.0 & 58.8 & 76.1 & 63.8 \\
\midrule
Bi-criteria \citep{wu2024active} & - & - & 68.94 & 78.25 & 87.96 \\
\bottomrule
\end{tabular}
} 
% \vspace{-0.1cm}
\caption{The Span-F1 scores of New Slot Unknown methods on five benchmarks.}
\label{tab:nsvd_main}
% \vspace{-0.5cm}
\end{table*}

\paragraph{New Slot Unknown.} Unlike the above studies, this setting focuses on extracting new slot values while also inducing potential new slots, without knowing the prior information of new slots. In this context, \citet{wu2022semi} used existing linguistic annotation tools to extract slot values and proposed an incremental clustering scheme that synergizes labeled and unlabeled data for slot structure discovery. To reduce labeling efforts with robust performance, \citet{wu2024active} introduced a Bi-criteria active learning scheme that selects data based on uncertainty and diversity when discerning new slots.

\subsection{Joint OnExp}
\label{tax:Joint ProOE}
While significant successes have been achieved, previous methods tackle new intent and slot-value discovery as separate tasks, despite their inherent interconnection. Joint OnExp addresses this by simultaneously identifying new intents, slots, and values, offering a comprehensive understanding but posing challenges in managing knowledge sharing without compromising performance. Pioneers in this field, \citet{joint_prooe_1} devised a coarse-to-fine three-step method—role-labeling, concept-mining, and pattern-mining—to infer intents, slots, and values. Despite its promising results, Joint OnExp is still under-explored, offering substantial space for further innovation.

\section{Leaderboard and Takeaway}

\paragraph{Leaderboard:} The leaderboard for representative NID and NSVD methods on widely recognized datasets is presented in Table \ref{tab:nid_main} and Table \ref{tab:nsvd_main}. More details are presented in Appendix \ref{app:leaderboard}.

\paragraph{Takeaway for NID:} Based on the review of NID efforts, we present the following observations:

\begin{itemize}[leftmargin=*, itemsep=0.1em]
    \item \textbf{\textit{Pre-trained Language Models Enhance OnExp.}} It has been observed that NID methods utilizing pre-trained models, such as CsePL and ALUP, consistently outperform traditional methods like PCK-Means by significant margins ($\sim$ 50\% in ACC). This demonstrates that pre-trained models, including LLMs, contribute substantial foundational knowledge and supplementary supervision signals. They enhance NID performance by offering a deeper contextual understanding and quicker adaptation to new user intents.
    \item \textbf{\textit{Prior Knowledge Leads to Improvement.}} We observe that NID methods with supervision generally surpass unsupervised ones, as incorporating prior knowledge—through labeled data or external information—significantly boosts the model's ability to identify new intents. For example, semi-supervised CsePL shows over 5\% improvements in all evaluation metrics compared to the SOTA unsupervised IDAS. This highlights the critical role of integrating prior knowledge.
\end{itemize}

\paragraph{Takeaway for NSVD:} According to the recent advances in NSVD, we have the following insights:

\begin{itemize}[leftmargin=*, itemsep=0.1em]
    \item \textbf{\textit{External Knowledge Enhances Results.}} Utilizing external knowledge bases in NSVD processes significantly enhances new slot value identification. These resources provide a rich contextual backdrop that aids models in accurately recognizing and categorizing new slot values, even in complex or ambiguous contexts.
    \item \textbf{\textit{Effective Knowledge Transfer Influences NSVD.}} Implementing effective knowledge transfer mechanisms that connect known slots and values with new slots and values enhances the ability of NSVD models. It leverages existing slot knowledge to inform and guide the identification and integration of new slots and values, reducing the learning curve and improving the system's adaptability to dynamic conversational contexts.
\end{itemize}

\section{Conclusion and Future Directions}

This paper presents the first comprehensive survey of recent advances in OnExp. We begin by formulating the task, detailing representative data resources and evaluation protocols used. We then examine prevalent OnExp methods, including NID, NSVD, and Joint OnExp. Despite significant progress achieved, several challenges remain, inspiring promising frontiers for future research.

\paragraph{Early OnExp.} 
Existing studies primarily concentrate on developing models to expand predefined ontologies using extensive utterances. Yet, real-world conversational agents necessitate the ability to rapidly recognize and adapt to evolving user needs and dialogue contexts \cite{li2023revisiting1,li2023revisiting2}, \nocite{li2024harnessing,li2022diaasq} thus highlighting the critical importance of early-stage OnExp. Early OnExp faces the unique challenge of identifying new ontological items with minimal utterances when a known ontology has been established using extensive data. In such a scenario, nascent ontological items risk being submerged by more prevalent ones. Although \citet{supnid_slm_13} showcased the effectiveness of CsePL in early intent discovery, more specific methods that fully address the unique challenges of this area remain largely under-explored. This highlights its significant potential as a promising field for future research.

\paragraph{Multi-modal OnExp.} 
Current OnExp tasks generally learned new ontological items from purely text-modal utterances. However, practical interactions with conversational agents typically occur in multi-modal settings \cite{liao2018knowledge,zhang2019neural,wu2022state}, suggesting that such multi-modal data can enhance new ontology learning. For example, incorporating visual data in e-commerce or audio cues in customer support could provide deeper contextual insights than text-only systems \citep{multimodal_1}. Despite its potential, multi-modal OnExp is still in its early stages, with limited research on effectively synergizing different modalities to expand ontologies. This emerging area promises to significantly improve the capabilities of conversational agents across different applications, necessitating more comprehensive research into advanced modality integration techniques and benchmarks of multi-modal data in OnExp.

\paragraph{Holistic OnExp.}
Prior OnExp research has mainly confined their ontology analyses to the CU module of conversational agents, assessing their performance via metrics such as recognition accuracy. This narrow focus, however, overlooks the broader impact of OnExp results on the other pivotal components of conversational agents, \textit{e.g.}, dialogue management and response generation. Additionally, the rationality of newly expanded ontologies has seldom been thoroughly examined, raising questions about whether OnExp outcomes can genuinely enhance dialogue policy learning or the quality of generated responses. To fill these gaps, there is a compelling need for more integrated approaches in OnExp. These methods should extend beyond merely identifying new ontological items, to a thorough evaluation of their holistic impact on the entire conversational agents, ensuring that advancements in OnExp positively contribute to the evolution of conversational AI and improve both system performance and user interaction quality.

% \section{Conclusion}
% This paper presents the first comprehensive survey of recent advances in OnExp, methodically categorizing the latest OnExp progress into three areas: NID, NSVD, and Joint OnExp. We provide detailed summaries of the representative data resources and evaluation protocols for each category. Further, the paper identifies several emerging frontiers in this field, aiming to advance research and innovation in OnExp. As a crucial step toward the evolution of conversational agents, it necessitates expanded research and deeper investigation into developing more advanced OnExp systems.

\section*{Limitations}
This survey provides a comprehensive overview of the latest studies in OnExp. Despite our diligent efforts, some limitations may still persist:

\paragraph{Categorization.} The survey makes the first attempt to organize the recent OnExp works into three distinct dimensions. This organization reflects our subjective interpretation and understanding. External insights on this categorization might enrich the perspectives presented.

\paragraph{Descriptions.} The descriptions of the introduced OnExp approaches in this survey are kept highly succinct to allow broad coverage within the constraints of page limits. We intend for this survey to act as a starting point, directing readers to the original works for more detailed information.

\paragraph{Experimental Results.} The leaderboard in this survey predominantly emphasizes broad comparisons of different OnExp approaches, such as the overarching system performance, instead of detailed analyses. Going forward, we aim to expand on these comparisons with more in-depth analyses of the experimental outcomes, thereby offering a more comprehensive understanding of the strengths and weaknesses of various OnExp models.

\section*{Acknowledgments}
This research is supported by the Ministry of Education, Singapore, under its AcRF Tier 2 Funding
(Proposal ID: T2EP20123-0052). Any opinions,
findings and conclusions or recommendations expressed in this material are those of the author(s)
and do not reflect the views of the Ministry of Education, Singapore.

\newpage
\bibliography{anthology,custom}

\begin{thebibliography}{121}
\expandafter\ifx\csname natexlab\endcsname\relax\def\natexlab#1{#1}\fi

\bibitem[{Aiello et~al.(2011)Aiello, Donato, Ozertem, and Menczer}]{unsupnid_stat_1}
Luca~Maria Aiello, Debora Donato, Umut Ozertem, and Filippo Menczer. 2011.
\newblock \href {https://doi.org/10.1145/2063576.2063775} {Behavior-driven clustering of queries into topics}.
\newblock In \emph{CIKM}, pages 1373--1382.

\bibitem[{An et~al.(2023{\natexlab{a}})An, Shi, Tian, Lin, Wang, Wu, Cai, Wang, Chen, Zhu, and Chen}]{supnid_hybrid_3}
Wenbin An, Wenkai Shi, Feng Tian, Haonan Lin, Qianying Wang, Yaqiang Wu, Mingxiang Cai, Luyan Wang, Yan Chen, Haiping Zhu, and Ping Chen. 2023{\natexlab{a}}.
\newblock \href {https://doi.org/10.48550/arXiv.2312.10897} {Generalized category discovery with large language models in the loop}.
\newblock \emph{CoRR}.

\bibitem[{An et~al.(2022)An, Tian, Chen, Tang, Zheng, and Wang}]{supnid_slm_15}
Wenbin An, Feng Tian, Ping Chen, Siliang Tang, Qinghua Zheng, and Qianying Wang. 2022.
\newblock \href {https://doi.org/10.18653/v1/2022.emnlp-main.85} {Fine-grained category discovery under coarse-grained supervision with hierarchical weighted self-contrastive learning}.
\newblock In \emph{EMNLP}, pages 1314--1323.

\bibitem[{An et~al.(2024)An, Tian, Shi, Chen, Wu, Wang, and Chen}]{supnid_slm_11}
Wenbin An, Feng Tian, Wenkai Shi, Yan Chen, Yaqiang Wu, Qianying Wang, and Ping Chen. 2024.
\newblock \href {https://doi.org/10.1609/aaai.v38i10.28959} {Transfer and alignment network for generalized category discovery}.
\newblock In \emph{AAAI}, pages 10856--10864.

\bibitem[{An et~al.(2023{\natexlab{b}})An, Tian, Zheng, Ding, Wang, and Chen}]{supnid_slm_10}
Wenbin An, Feng Tian, Qinghua Zheng, Wei Ding, Qianying Wang, and Ping Chen. 2023{\natexlab{b}}.
\newblock \href {https://doi.org/10.1609/aaai.v37i11.26475} {Generalized category discovery with decoupled prototypical network}.
\newblock In \emph{AAAI}, pages 12527--12535.

\bibitem[{Anand et~al.(2023)Anand, Venktesh, Anand, and Setty}]{Anand2023QueryUI}
Avishek Anand, V.~Venktesh, Abhijit Anand, and Vinay Setty. 2023.
\newblock \href {https://api.semanticscholar.org/CorpusID:259274869} {Query understanding in the age of large language models}.
\newblock \emph{ArXiv}.

\bibitem[{Bapna et~al.(2017)Bapna, T{\"{u}}r, Hakkani{-}T{\"{u}}r, and Heck}]{semi_nvd_based_1}
Ankur Bapna, G{\"{o}}khan T{\"{u}}r, Dilek Hakkani{-}T{\"{u}}r, and Larry~P. Heck. 2017.
\newblock \href {https://doi.org/10.21437/Interspeech.2017-518} {Towards zero-shot frame semantic parsing for domain scaling}.
\newblock In \emph{INTERSPEECH}, pages 2476--2480.

\bibitem[{Basu et~al.(2004)Basu, Banerjee, and Mooney}]{supnid_slm_1}
Sugato Basu, Arindam Banerjee, and Raymond~J. Mooney. 2004.
\newblock \href {https://doi.org/10.1137/1.9781611972740.31} {Active semi-supervision for pairwise constrained clustering}.
\newblock In \emph{ICDM}, pages 333--344.

\bibitem[{Budzianowski et~al.(2018)Budzianowski, Wen, Tseng, Casanueva, Ultes, Ramadan, and Ga{\v{s}}i{\'c}}]{budzianowski-etal-2018-multiwoz-1}
Pawe{\l} Budzianowski, Tsung-Hsien Wen, Bo-Hsiang Tseng, I{\~n}igo Casanueva, Stefan Ultes, Osman Ramadan, and Milica Ga{\v{s}}i{\'c}. 2018.
\newblock \href {https://aclanthology.org/D18-1547} {{M}ulti{WOZ} - a large-scale multi-domain {W}izard-of-{O}z dataset for task-oriented dialogue modelling}.
\newblock In \emph{EMNLP}, pages 5016--5026.

\bibitem[{Carmel et~al.(2018)Carmel, Lewin{-}Eytan, and Maarek}]{Alexa}
David Carmel, Liane Lewin{-}Eytan, and Yoelle Maarek. 2018.
\newblock \href {https://doi.org/10.1145/3209978.3210203} {Product question answering using customer generated content - research challenges}.
\newblock In \emph{SIGIR}, pages 1349--1350.

\bibitem[{Caron et~al.(2018)Caron, Bojanowski, Joulin, and Douze}]{unsupnid_nn_4}
Mathilde Caron, Piotr Bojanowski, Armand Joulin, and Matthijs Douze. 2018.
\newblock \href {https://doi.org/10.1007/978-3-030-01264-9\_9} {Deep clustering for unsupervised learning of visual features}.
\newblock In \emph{ECCV}, pages 139--156.

\bibitem[{Casanueva et~al.(2020)Casanueva, Tem{\v{c}}inas, Gerz, Henderson, and Vuli{\'c}}]{banking_dataset}
I{\~n}igo Casanueva, Tadas Tem{\v{c}}inas, Daniela Gerz, Matthew Henderson, and Ivan Vuli{\'c}. 2020.
\newblock \href {https://aclanthology.org/2020.nlp4convai-1.5} {Efficient intent detection with dual sentence encoders}.
\newblock In \emph{NLP4ConvAI@ACL}, pages 38--45.

\bibitem[{Chang et~al.(2017)Chang, Wang, Meng, Xiang, and Pan}]{unsupnid_nn_3}
Jianlong Chang, Lingfeng Wang, Gaofeng Meng, Shiming Xiang, and Chunhong Pan. 2017.
\newblock \href {https://doi.org/10.1109/ICCV.2017.626} {Deep adaptive image clustering}.
\newblock In \emph{ICCV}, pages 5880--5888.

\bibitem[{Chen and Moschitti(2019)}]{nvd_based_8}
Lingzhen Chen and Alessandro Moschitti. 2019.
\newblock \href {https://doi.org/10.1609/aaai.v33i01.33016260} {Transfer learning for sequence labeling using source model and target data}.
\newblock In \emph{AAAI}, pages 6260--6267.

\bibitem[{Chen et~al.(2015)Chen, Wang, and Rudnicky}]{chen2015jointly}
Yun-Nung Chen, William~Yang Wang, and Alexander Rudnicky. 2015.
\newblock Jointly modeling inter-slot relations by random walk on knowledge graphs for unsupervised spoken language understanding.
\newblock In \emph{NAACL}, pages 619--629.

\bibitem[{Chen et~al.(2013)Chen, Wang, and Rudnicky}]{Unsupervised_induction}
Yun-Nung Chen, William~Yang Wang, and Alexander~I. Rudnicky. 2013.
\newblock Unsupervised induction and filling of semantic slots for spoken dialogue systems using frame-semantic parsing.
\newblock In \emph{2013 IEEE Workshop on Automatic Speech Recognition and Understanding}, pages 120--125.

\bibitem[{Chen et~al.(2014)Chen, Wang, and Rudnicky}]{chen2014leveraging}
Yun-Nung Chen, William~Yang Wang, and Alexander~I Rudnicky. 2014.
\newblock Leveraging frame semantics and distributional semantics for unsupervised semantic slot induction in spoken dialogue systems.
\newblock In \emph{SLT}, pages 584--589.

\bibitem[{Cheung and Li(2012)}]{unsupnid_stat_2}
Jackie Chi~Kit Cheung and Xiao Li. 2012.
\newblock \href {https://doi.org/10.1145/2124295.2124342} {Sequence clustering and labeling for unsupervised query intent discovery}.
\newblock In \emph{WSDM}, pages 383--392.

\bibitem[{Coope et~al.(2020)Coope, Farghly, Gerz, Vulic, and Henderson}]{nvd_based_5}
Sam Coope, Tyler Farghly, Daniela Gerz, Ivan Vulic, and Matthew Henderson. 2020.
\newblock \href {https://doi.org/10.18653/v1/2020.acl-main.11} {Span-convert: Few-shot span extraction for dialog with pretrained conversational representations}.
\newblock In \emph{ACL}, pages 107--121.

\bibitem[{Coucke et~al.(2018)Coucke, Saade, Ball, Bluche, Caulier, Leroy, Doumouro, Gisselbrecht, Caltagirone, Lavril, Primet, and Dureau}]{Snips_dataset}
Alice Coucke, Alaa Saade, Adrien Ball, Th{\'{e}}odore Bluche, Alexandre Caulier, David Leroy, Cl{\'{e}}ment Doumouro, Thibault Gisselbrecht, Francesco Caltagirone, Thibaut Lavril, Ma{\"{e}}l Primet, and Joseph Dureau. 2018.
\newblock \href {http://arxiv.org/abs/1805.10190} {Snips voice platform: an embedded spoken language understanding system for private-by-design voice interfaces}.
\newblock \emph{CoRR}.

\bibitem[{Dao et~al.(2024)Dao, Deng, Le, and Liao}]{DBLP:conf/sigir/Dao0LL24}
Huy Dao, Yang Deng, Dung~D. Le, and Lizi Liao. 2024.
\newblock \href {https://doi.org/10.1145/3626772.3657755} {Broadening the view: Demonstration-augmented prompt learning for conversational recommendation}.
\newblock In \emph{SIGIR}, pages 785--795.

\bibitem[{Dao et~al.(2023)Dao, Liao, Le, and Nie}]{dao-etal-2023-reinforced}
Huy Dao, Lizi Liao, Dung Le, and Yuxiang Nie. 2023.
\newblock \href {https://aclanthology.org/2023.emnlp-main.775} {Reinforced target-driven conversational promotion}.
\newblock In \emph{EMNLP}, pages 12583--12596.

\bibitem[{De~Raedt et~al.(2023)De~Raedt, Godin, Demeester, and Develder}]{unsupnid_nn_6}
Maarten De~Raedt, Fr{\'e}deric Godin, Thomas Demeester, and Chris Develder. 2023.
\newblock \href {https://aclanthology.org/2023.nlp4convai-1.7} {{IDAS}: Intent discovery with abstractive summarization}.
\newblock In \emph{NLP4ConvAI@ACL}, pages 71--88.

\bibitem[{Devlin et~al.(2019)Devlin, Chang, Lee, and Toutanova}]{bert/DevlinCLT19}
Jacob Devlin, Ming{-}Wei Chang, Kenton Lee, and Kristina Toutanova. 2019.
\newblock \href {https://doi.org/10.18653/v1/n19-1423} {{BERT:} pre-training of deep bidirectional transformers for language understanding}.
\newblock In \emph{NAACL-HLT}, pages 4171--4186.

\bibitem[{Du et~al.(2021)Du, He, Li, Yu, Pasupat, and Zhang}]{QASF}
Xinya Du, Luheng He, Qi~Li, Dian Yu, Panupong Pasupat, and Yuan Zhang. 2021.
\newblock \href {https://doi.org/10.18653/v1/2021.acl-short.83} {Qa-driven zero-shot slot filling with weak supervision pretraining}.
\newblock In \emph{ACL/IJCNLP}, pages 654--664.

\bibitem[{E et~al.(2019)E, Niu, Chen, and Song}]{intent_dec_1}
Haihong E, Peiqing Niu, Zhongfu Chen, and Meina Song. 2019.
\newblock \href {https://aclanthology.org/P19-1544} {A novel bi-directional interrelated model for joint intent detection and slot filling}.
\newblock In \emph{ACL}, pages 5467--5471.

\bibitem[{Eric et~al.(2020)Eric, Goel, Paul, Sethi, Agarwal, Gao, Kumar, Goyal, Ku, and Hakkani{-}T{\"{u}}r}]{multiwoz/EricGPSAGKGKH20}
Mihail Eric, Rahul Goel, Shachi Paul, Abhishek Sethi, Sanchit Agarwal, Shuyang Gao, Adarsh Kumar, Anuj~Kumar Goyal, Peter Ku, and Dilek Hakkani{-}T{\"{u}}r. 2020.
\newblock \href {https://aclanthology.org/2020.lrec-1.53/} {Multiwoz 2.1: {A} consolidated multi-domain dialogue dataset with state corrections and state tracking baselines}.
\newblock In \emph{LREC}, pages 422--428.

\bibitem[{Gowda and Krishna(1978)}]{unsupnid_stat_7}
K.~Chidananda Gowda and G.~Krishna. 1978.
\newblock \href {https://doi.org/10.1016/0031-3203(78)90018-3} {Agglomerative clustering using the concept of mutual nearest neighbourhood}.
\newblock \emph{Pattern Recognit.}, pages 105--112.

\bibitem[{Guo et~al.(2024)Guo, Liao, Zhang, Wang, Li, and Chen}]{DBLP:conf/naacl/GuoLZWLC24}
Shasha Guo, Lizi Liao, Jing Zhang, Yanling Wang, Cuiping Li, and Hong Chen. 2024.
\newblock \href {https://doi.org/10.18653/v1/2024.findings-naacl.287} {{SGSH:} stimulate large language models with skeleton heuristics for knowledge base question generation}.
\newblock In \emph{Findings of NAACL}, pages 4613--4625.

\bibitem[{Hakkani{-}T{\"{u}}r et~al.(2013)Hakkani{-}T{\"{u}}r, Celikyilmaz, Heck, and T{\"{u}}r}]{unsupnid_stat_4}
Dilek Hakkani{-}T{\"{u}}r, Asli Celikyilmaz, Larry~P. Heck, and G{\"{o}}khan T{\"{u}}r. 2013.
\newblock \href {https://doi.org/10.21437/Interspeech.2013-598} {A weakly-supervised approach for discovering new user intents from search query logs}.
\newblock In \emph{INTERSPEECH}, pages 3780--3784.

\bibitem[{Hakkani{-}T{\"{u}}r et~al.(2015)Hakkani{-}T{\"{u}}r, Ju, Zweig, and T{\"{u}}r}]{unsupnid_stat_5}
Dilek Hakkani{-}T{\"{u}}r, Yun{-}Cheng Ju, Geoffrey Zweig, and G{\"{o}}khan T{\"{u}}r. 2015.
\newblock \href {https://doi.org/10.21437/Interspeech.2015-70} {Clustering novel intents in a conversational interaction system with semantic parsing}.
\newblock In \emph{INTERSPEECH}, pages 1854--1858.

\bibitem[{Han et~al.(2019)Han, Vedaldi, and Zisserman}]{supnid_slm_4}
Kai Han, Andrea Vedaldi, and Andrew Zisserman. 2019.
\newblock \href {https://doi.org/10.1109/ICCV.2019.00849} {Learning to discover novel visual categories via deep transfer clustering}.
\newblock In \emph{ICCV}, pages 8400--8408.

\bibitem[{He et~al.(2020{\natexlab{a}})He, Yan, Xu, Liu, Liu, and Xu}]{nvd_based_9}
Keqing He, Yuanmeng Yan, Hong Xu, Sihong Liu, Zijun Liu, and Weiran Xu. 2020{\natexlab{a}}.
\newblock \href {https://doi.org/10.1109/IJCNN48605.2020.9206988} {Learning label-relational output structure for adaptive sequence labeling}.
\newblock In \emph{IJCNN}, pages 1--8.

\bibitem[{He et~al.(2020{\natexlab{b}})He, Yan, and Xu}]{nvd_based_4}
Keqing He, Yuanmeng Yan, and Weiran Xu. 2020{\natexlab{b}}.
\newblock \href {https://doi.org/10.18653/v1/2020.acl-main.58} {Learning to tag {OOV} tokens by integrating contextual representation and background knowledge}.
\newblock In \emph{ACL}, pages 619--624.

\bibitem[{He et~al.(2020{\natexlab{c}})He, Zhang, Yan, Xu, Niu, and Zhou}]{CZSL-Adv}
Keqing He, Jinchao Zhang, Yuanmeng Yan, Weiran Xu, Cheng Niu, and Jie Zhou. 2020{\natexlab{c}}.
\newblock \href {https://aclanthology.org/2020.coling-main.126} {Contrastive zero-shot learning for cross-domain slot filling with adversarial attack}.
\newblock In \emph{ACL}, pages 1461--1467.

\bibitem[{Hemphill et~al.(1990)Hemphill, Godfrey, and Doddington}]{atis_dataset}
Charles~T. Hemphill, John~J. Godfrey, and George~R. Doddington. 1990.
\newblock \href {https://aclanthology.org/H90-1021/} {The {ATIS} spoken language systems pilot corpus}.
\newblock In \emph{Speech and Natural Language: Workshop}.

\bibitem[{Henderson et~al.(2012)Henderson, Gasic, Thomson, Tsiakoulis, Yu, and Young}]{canbridge_slu_dataset}
Matthew Henderson, Milica Gasic, Blaise Thomson, Pirros Tsiakoulis, Kai Yu, and Steve~J. Young. 2012.
\newblock \href {https://doi.org/10.1109/SLT.2012.6424218} {Discriminative spoken language understanding using word confusion networks}.
\newblock In \emph{SLT}, pages 176--181.

\bibitem[{Hou et~al.(2020)Hou, Che, Lai, Zhou, Liu, Liu, and Liu}]{L-TapNet/acl/HouCLZLLL20}
Yutai Hou, Wanxiang Che, Yongkui Lai, Zhihan Zhou, Yijia Liu, Han Liu, and Ting Liu. 2020.
\newblock \href {https://doi.org/10.18653/v1/2020.acl-main.128} {Few-shot slot tagging with collapsed dependency transfer and label-enhanced task-adaptive projection network}.
\newblock In \emph{ACL}, pages 1381--1393.

\bibitem[{Hsu et~al.(2018)Hsu, Lv, and Kira}]{supnid_slm_2}
Yen{-}Chang Hsu, Zhaoyang Lv, and Zsolt Kira. 2018.
\newblock \href {https://openreview.net/forum?id=ByRWCqvT-} {Learning to cluster in order to transfer across domains and tasks}.
\newblock In \emph{ICLR}.

\bibitem[{Hsu et~al.(2019)Hsu, Lv, Schlosser, Odom, and Kira}]{supnid_slm_3}
Yen{-}Chang Hsu, Zhaoyang Lv, Joel Schlosser, Phillip Odom, and Zsolt Kira. 2019.
\newblock \href {https://openreview.net/forum?id=SJzR2iRcK7} {Multi-class classification without multi-class labels}.
\newblock In \emph{ICLR}.

\bibitem[{Hu et~al.(2019)Hu, Chen, Chang, and Sun}]{nvd_based_6}
Ziniu Hu, Ting Chen, Kai-Wei Chang, and Yizhou Sun. 2019.
\newblock \href {https://aclanthology.org/P19-1402} {Few-shot representation learning for out-of-vocabulary words}.
\newblock In \emph{ACL}, pages 4102--4112.

\bibitem[{Hude{\v{c}}ek et~al.(2021)Hude{\v{c}}ek, Du{\v{s}}ek, and Yu}]{hudevcek2021discovering}
Vojt{\v{e}}ch Hude{\v{c}}ek, Ond{\v{r}}ej Du{\v{s}}ek, and Zhou Yu. 2021.
\newblock Discovering dialogue slots with weak supervision.
\newblock In \emph{ACL-IJCNLP}, pages 2430--2442.

\bibitem[{Jansen et~al.(2008)Jansen, Booth, and Spink}]{unsupnid_rule_2}
Bernard~J. Jansen, Danielle~L. Booth, and Amanda Spink. 2008.
\newblock \href {https://doi.org/10.1016/j.ipm.2007.07.015} {Determining the informational, navigational, and transactional intent of web queries}.
\newblock \emph{Inf. Process. Manag.}, pages 1251--1266.

\bibitem[{Kumar et~al.(2022)Kumar, Patidar, Varshney, Vig, and Shroff}]{supnid_slm_7.5}
Rajat Kumar, Mayur Patidar, Vaibhav Varshney, Lovekesh Vig, and Gautam Shroff. 2022.
\newblock \href {https://doi.org/10.18653/v1/2022.naacl-main.134} {Intent detection and discovery from user logs via deep semi-supervised contrastive clustering}.
\newblock In \emph{NAACL-HLT}, pages 1836--1853.

\bibitem[{Lamanov et~al.(2022)Lamanov, Burnyshev, Artemova, Malykh, Bout, and Piontkovskaya}]{lamanov-2022-template}
Dmitry Lamanov, Pavel Burnyshev, Ekaterina Artemova, Valentin Malykh, Andrey Bout, and Irina Piontkovskaya. 2022.
\newblock \href {https://aclanthology.org/2022.inlg-main.2} {Template-based approach to zero-shot intent recognition}.
\newblock In \emph{INLG}.

\bibitem[{Larson and Leach(2022)}]{DBLP:journals/corr/abs-2207-13211}
Stefan Larson and Kevin Leach. 2022.
\newblock \href {https://doi.org/10.48550/arXiv.2207.13211} {A survey of intent classification and slot-filling datasets for task-oriented dialog}.
\newblock \emph{CoRR}, abs/2207.13211.

\bibitem[{Larson et~al.(2019)Larson, Mahendran, Peper, Clarke, Lee, Hill, Kummerfeld, Leach, Laurenzano, Tang, and Mars}]{clinc_dataset}
Stefan Larson, Anish Mahendran, Joseph~J. Peper, Christopher Clarke, Andrew Lee, Parker Hill, Jonathan~K. Kummerfeld, Kevin Leach, Michael~A. Laurenzano, Lingjia Tang, and Jason Mars. 2019.
\newblock \href {https://aclanthology.org/D19-1131} {An evaluation dataset for intent classification and out-of-scope prediction}.
\newblock In \emph{EMNLP-IJCNLP}, pages 1311--1316.

\bibitem[{Lee and Jha(2019)}]{zat/aaai/LeeJ19}
Sungjin Lee and Rahul Jha. 2019.
\newblock \href {https://doi.org/10.1609/aaai.v33i01.33016642} {Zero-shot adaptive transfer for conversational language understanding}.
\newblock In \emph{AAAI}, pages 6642--6649.

\bibitem[{Lewis et~al.(2020)Lewis, Liu, Goyal, Ghazvininejad, Mohamed, Levy, Stoyanov, and Zettlemoyer}]{bart/LewisLGGMLSZ20}
Mike Lewis, Yinhan Liu, Naman Goyal, Marjan Ghazvininejad, Abdelrahman Mohamed, Omer Levy, Veselin Stoyanov, and Luke Zettlemoyer. 2020.
\newblock \href {https://doi.org/10.18653/v1/2020.acl-main.703} {{BART:} denoising sequence-to-sequence pre-training for natural language generation, translation, and comprehension}.
\newblock In \emph{ACL}, pages 7871--7880.

\bibitem[{Li et~al.(2023{\natexlab{a}})Li, Fei, Li, Wu, Liao, Wei, Chua, and Ji}]{li2023revisiting1}
Bobo Li, Hao Fei, Fei Li, Shengqiong Wu, Lizi Liao, Yinwei Wei, Tat-Seng Chua, and Donghong Ji. 2023{\natexlab{a}}.
\newblock Revisiting conversation discourse for dialogue disentanglement.
\newblock \emph{ACM Transactions on Information Systems (TOIS)}.

\bibitem[{Li et~al.(2022)Li, Fei, Li, Wu, Zhang, Wu, Li, Liu, Liao, Chua et~al.}]{li2022diaasq}
Bobo Li, Hao Fei, Fei Li, Yuhan Wu, Jinsong Zhang, Shengqiong Wu, Jingye Li, Yijiang Liu, Lizi Liao, Tat-Seng Chua, et~al. 2022.
\newblock Diaasq: A benchmark of conversational aspect-based sentiment quadruple analysis.
\newblock In \emph{ACL}.

\bibitem[{Li et~al.(2024)Li, Fei, Liao, Zhao, Su, Li, and Ji}]{li2024harnessing}
Bobo Li, Hao Fei, Lizi Liao, Yu~Zhao, Fangfang Su, Fei Li, and Donghong Ji. 2024.
\newblock Harnessing holistic discourse features and triadic interaction for sentiment quadruple extraction in dialogues.
\newblock In \emph{AAAI}, pages 18462--18470.

\bibitem[{Li et~al.(2023{\natexlab{b}})Li, Fei, Liao, Zhao, Teng, Chua, Ji, and Li}]{li2023revisiting2}
Bobo Li, Hao Fei, Lizi Liao, Yu~Zhao, Chong Teng, Tat-Seng Chua, Donghong Ji, and Fei Li. 2023{\natexlab{b}}.
\newblock Revisiting disentanglement and fusion on modality and context in conversational multimodal emotion recognition.
\newblock In \emph{ACM MM}, pages 5923--5934.

\bibitem[{Li et~al.(2017)Li, Qiu, Chen, Wang, Gao, Huang, Ren, Zhao, Zhao, Wang, Jin, and Chu}]{AliMe}
Feng{-}Lin Li, Minghui Qiu, Haiqing Chen, Xiongwei Wang, Xing Gao, Jun Huang, Juwei Ren, Zhongzhou Zhao, Weipeng Zhao, Lei Wang, Guwei Jin, and Wei Chu. 2017.
\newblock \href {https://doi.org/10.1145/3132847.3133169} {\emph{AliMe Assist }: An intelligent assistant for creating an innovative e-commerce experience}.
\newblock In \emph{CIKM}, pages 2495--2498.

\bibitem[{Li et~al.(2023{\natexlab{c}})Li, Wang, Dong, He, Zhao, Lei, Liu, and Xu}]{GZPL}
Xuefeng Li, Liwen Wang, Guanting Dong, Keqing He, Jinzheng Zhao, Hao Lei, Jiachi Liu, and Weiran Xu. 2023{\natexlab{c}}.
\newblock \href {https://doi.org/10.18653/v1/2023.findings-acl.52} {Generative zero-shot prompt learning for cross-domain slot filling with inverse prompting}.
\newblock In \emph{Findings of ACL}, pages 825--834.

\bibitem[{Liang et~al.(2017)Liang, Xu, and Zhao}]{nvd_based_1}
Dongyun Liang, Weiran Xu, and Yinge Zhao. 2017.
\newblock \href {https://doi.org/10.18653/v1/w17-2606} {Combining word-level and character-level representations for relation classification of informal text}.
\newblock In \emph{Rep4NLP@ACL}, pages 43--47.

\bibitem[{Liang and Liao(2023)}]{supnid_slm_13}
Jinggui Liang and Lizi Liao. 2023.
\newblock \href {https://aclanthology.org/2023.findings-emnlp.702} {Clusterprompt: Cluster semantic enhanced prompt learning for new intent discovery}.
\newblock In \emph{Findings of EMNLP}, pages 10468--10481.

\bibitem[{Liang et~al.(2024{\natexlab{a}})Liang, Liao, Fei, and Jiang}]{liang-etal-2024-synergizing}
Jinggui Liang, Lizi Liao, Hao Fei, and Jing Jiang. 2024{\natexlab{a}}.
\newblock \href {https://doi.org/10.18653/v1/2024.findings-acl.840} {Synergizing large language models and pre-trained smaller models for conversational intent discovery}.
\newblock In \emph{Findings of the Association for Computational Linguistics ACL 2024}, pages 14133--14147.

\bibitem[{Liang et~al.(2024{\natexlab{b}})Liang, Liao, Fei, Li, and Jiang}]{supnid_hybrid_4}
Jinggui Liang, Lizi Liao, Hao Fei, Bobo Li, and Jing Jiang. 2024{\natexlab{b}}.
\newblock \href {https://api.semanticscholar.org/CorpusID:268880579} {Actively learn from llms with uncertainty propagation for generalized category discovery}.
\newblock In \emph{NAACL-HLT}.

\bibitem[{Liao et~al.(2018)Liao, Ma, He, Hong, and Chua}]{liao2018knowledge}
Lizi Liao, Yunshan Ma, Xiangnan He, Richang Hong, and Tat-seng Chua. 2018.
\newblock Knowledge-aware multimodal dialogue systems.
\newblock In \emph{Proceedings of the 26th ACM international conference on Multimedia}, pages 801--809.

\bibitem[{Lin et~al.(2020)Lin, Xu, and Zhang}]{supnid_slm_5}
Ting{-}En Lin, Hua Xu, and Hanlei Zhang. 2020.
\newblock \href {https://doi.org/10.1609/aaai.v34i05.6353} {Discovering new intents via constrained deep adaptive clustering with cluster refinement}.
\newblock In \emph{AAAI}, pages 8360--8367.

\bibitem[{Liu et~al.(2019)Liu, Zhang, Fan, Fu, Li, Wu, and Lam}]{emnlp/LiuZFFLWL19}
Han Liu, Xiaotong Zhang, Lu~Fan, Xuandi Fu, Qimai Li, Xiao{-}Ming Wu, and Albert Y.~S. Lam. 2019.
\newblock \href {https://doi.org/10.18653/v1/D19-1486} {Reconstructing capsule networks for zero-shot intent classification}.
\newblock In \emph{EMNLP-IJCNLP}, pages 4798--4808.

\bibitem[{Liu et~al.(2022{\natexlab{a}})Liu, Zhao, Zhang, Zhang, Sun, Yu, and Zhang}]{sigir/LiuZZZS0Z22}
Han Liu, Siyang Zhao, Xiaotong Zhang, Feng Zhang, Junjie Sun, Hong Yu, and Xianchao Zhang. 2022{\natexlab{a}}.
\newblock \href {https://doi.org/10.1145/3477495.3531803} {A simple meta-learning paradigm for zero-shot intent classification with mixture attention mechanism}.
\newblock In \emph{SIGIR}, pages 2047--2052.

\bibitem[{Liu et~al.(2022{\natexlab{b}})Liu, Yu, Chen, and Xu}]{RCSF}
Jian Liu, Mengshi Yu, Yufeng Chen, and Jinan Xu. 2022{\natexlab{b}}.
\newblock \href {https://doi.org/10.1109/TASLP.2022.3140559} {Cross-domain slot filling as machine reading comprehension: {A} new perspective}.
\newblock \emph{{IEEE} {ACM} Trans. Audio Speech Lang. Process.}, pages 673--685.

\bibitem[{Liu et~al.(2023)Liu, Fabbri, Liu, Radev, and Cohan}]{Liu2023OnLT}
Yixin Liu, Alexander~R. Fabbri, Pengfei Liu, Dragomir~R. Radev, and Arman Cohan. 2023.
\newblock \href {https://api.semanticscholar.org/CorpusID:258841126} {On learning to summarize with large language models as references}.
\newblock \emph{ArXiv}.

\bibitem[{Liu et~al.(2020)Liu, Winata, Xu, and Fung}]{Coach}
Zihan Liu, Genta~Indra Winata, Peng Xu, and Pascale Fung. 2020.
\newblock \href {https://doi.org/10.18653/v1/2020.acl-main.3} {Coach: {A} coarse-to-fine approach for cross-domain slot filling}.
\newblock In \emph{ACL}, pages 19--25.

\bibitem[{Luo and Liu(2023)}]{SP-prompting}
Qiaoyang Luo and Lingqiao Liu. 2023.
\newblock \href {https://api.semanticscholar.org/CorpusID:259631454} {Zero-shot slot filling with slot-prefix prompting and attention relationship descriptor}.
\newblock In \emph{AAAI}.

\bibitem[{MacQueen et~al.(1967)}]{unsupnid_stat_6}
James MacQueen et~al. 1967.
\newblock Some methods for classification and analysis of multivariate observations.
\newblock In \emph{Proceedings of the fifth Berkeley symposium on mathematical statistics and probability}, pages 281--297.

\bibitem[{Mou et~al.(2022{\natexlab{a}})Mou, He, Wang, Wu, Wang, Wu, and Xu}]{supnid_slm_7.7}
Yutao Mou, Keqing He, Pei Wang, Yanan Wu, Jingang Wang, Wei Wu, and Weiran Xu. 2022{\natexlab{a}}.
\newblock \href {https://doi.org/10.18653/v1/2022.emnlp-main.98} {Watch the neighbors: {A} unified k-nearest neighbor contrastive learning framework for {OOD} intent discovery}.
\newblock In \emph{EMNLP}, pages 1517--1529.

\bibitem[{Mou et~al.(2022{\natexlab{b}})Mou, He, Wu, Zeng, Xu, Jiang, Wu, and Xu}]{supnid_slm_7.8}
Yutao Mou, Keqing He, Yanan Wu, Zhiyuan Zeng, Hong Xu, Huixing Jiang, Wei Wu, and Weiran Xu. 2022{\natexlab{b}}.
\newblock \href {https://doi.org/10.18653/v1/2022.acl-short.6} {Disentangled knowledge transfer for {OOD} intent discovery with unified contrastive learning}.
\newblock In \emph{ACL}, pages 46--53.

\bibitem[{Mou et~al.(2023)Mou, Song, He, Zeng, Wang, Wang, Xian, and Xu}]{supnid_slm_9.5}
Yutao Mou, Xiaoshuai Song, Keqing He, Chen Zeng, Pei Wang, Jingang Wang, Yunsen Xian, and Weiran Xu. 2023.
\newblock \href {https://doi.org/10.18653/v1/2023.acl-long.538} {Decoupling pseudo label disambiguation and representation learning for generalized intent discovery}.
\newblock In \emph{ACL}, pages 9661--9675.

\bibitem[{Mrksic et~al.(2017)Mrksic, S{\'{e}}aghdha, Wen, Thomson, and Young}]{ontology}
Nikola Mrksic, Diarmuid~{\'{O}} S{\'{e}}aghdha, Tsung{-}Hsien Wen, Blaise Thomson, and Steve~J. Young. 2017.
\newblock \href {https://doi.org/10.18653/v1/P17-1163} {Neural belief tracker: Data-driven dialogue state tracking}.
\newblock In \emph{ACL}, pages 1777--1788.

\bibitem[{Neves~Ribeiro et~al.(2023)Neves~Ribeiro, Goetz, Abdar, Ross, Dong, Forbus, and Mohamed}]{ontology-2}
Danilo Neves~Ribeiro, Jack Goetz, Omid Abdar, Mike Ross, Annie Dong, Kenneth Forbus, and Ahmed Mohamed. 2023.
\newblock \href {https://aclanthology.org/2023.pandl-1.6} {Towards zero-shot frame semantic parsing with task agnostic ontologies and simple labels}.
\newblock In \emph{Proceedings of the 2nd Workshop on Pattern-based Approaches to NLP in the Age of Deep Learning}, pages 54--63.

\bibitem[{Nguyen et~al.(2023)Nguyen, Zhang, Liu, and Yu}]{nguyen-etal-2023-slot}
Hoang Nguyen, Chenwei Zhang, Ye~Liu, and Philip Yu. 2023.
\newblock \href {https://aclanthology.org/2023.sigdial-1.44} {Slot induction via pre-trained language model probing and multi-level contrastive learning}.
\newblock In \emph{Proceedings of the 24th Meeting of the Special Interest Group on Discourse and Dialogue}, pages 470--481, Prague, Czechia. Association for Computational Linguistics.

\bibitem[{Oguz and Vu(2021)}]{ARN/eacl/OguzV21}
Cennet Oguz and Ngoc~Thang Vu. 2021.
\newblock \href {https://doi.org/10.18653/v1/2021.eacl-main.134} {Few-shot learning for slot tagging with attentive relational network}.
\newblock In \emph{EACL}, pages 1566--1572.

\bibitem[{OpenAI(2023)}]{Achiam2023GPT4TR}
OpenAI. 2023.
\newblock \href {https://doi.org/10.48550/arXiv.2303.08774} {{GPT-4} technical report}.
\newblock \emph{CoRR}, abs/2303.08774.

\bibitem[{Parikh et~al.(2023)Parikh, Tiwari, Tumbade, and Vohra}]{acl/ParikhTTV23}
Soham Parikh, Mitul Tiwari, Prashil Tumbade, and Quaizar Vohra. 2023.
\newblock \href {https://doi.org/10.18653/v1/2023.acl-industry.71} {Exploring zero and few-shot techniques for intent classification}.
\newblock In \emph{ACL}, pages 744--751.

\bibitem[{Rastogi et~al.(2020)Rastogi, Zang, Sunkara, Gupta, and Khaitan}]{SGD/RastogiZSGK20}
Abhinav Rastogi, Xiaoxue Zang, Srinivas Sunkara, Raghav Gupta, and Pranav Khaitan. 2020.
\newblock \href {https://doi.org/10.1609/aaai.v34i05.6394} {Towards scalable multi-domain conversational agents: The schema-guided dialogue dataset}.
\newblock In \emph{AAAI}, pages 8689--8696.

\bibitem[{Ren et~al.(2014)Ren, Wang, Yu, Yan, Chen, and Han}]{unsupnid_stat_3}
Xiang Ren, Yujing Wang, Xiao Yu, Jun Yan, Zheng Chen, and Jiawei Han. 2014.
\newblock \href {https://doi.org/10.1145/2556195.2556222} {Heterogeneous graph-based intent learning with queries, web pages and wikipedia concepts}.
\newblock In \emph{WSDM}, pages 23--32.

\bibitem[{Rose and Levinson(2004)}]{unsupnid_rule_1}
Daniel~E. Rose and Danny Levinson. 2004.
\newblock \href {https://doi.org/10.1145/988672.988675} {Understanding user goals in web search}.
\newblock In \emph{WWW}, pages 13--19.

\bibitem[{Shah et~al.(2019)Shah, Gupta, Fayazi, and Hakkani{-}T{\"{u}}r}]{nvd_based_7}
Darsh~J. Shah, Raghav Gupta, Amir~A. Fayazi, and Dilek Hakkani{-}T{\"{u}}r. 2019.
\newblock \href {https://doi.org/10.18653/v1/p19-1547} {Robust zero-shot cross-domain slot filling with example values}.
\newblock In \emph{ACL}, pages 5484--5490.

\bibitem[{Shen et~al.(2021)Shen, Sun, Zhang, and Najmabadi}]{supnid_slm_7}
Xiang Shen, Yinge Sun, Yao Zhang, and Mani Najmabadi. 2021.
\newblock Semi-supervised intent discovery with contrastive learning.
\newblock In \emph{NLP4CONVAI}, pages 120--129.

\bibitem[{Shi et~al.(2023)Shi, An, Tian, Zheng, Wang, and Chen}]{supnid_slm_14}
Wenkai Shi, Wenbin An, Feng Tian, Qinghua Zheng, Qianying Wang, and Ping Chen. 2023.
\newblock \href {https://aclanthology.org/2023.emnlp-main.499} {A diffusion weighted graph framework for new intent discovery}.
\newblock In \emph{EMNLP}, pages 8033--8042.

\bibitem[{Si et~al.(2021)Si, Liu, Fu, Lin, Li, and Wang}]{ijcai/SiL00LW21}
Qingyi Si, Yuanxin Liu, Peng Fu, Zheng Lin, Jiangnan Li, and Weiping Wang. 2021.
\newblock \href {https://doi.org/10.24963/ijcai.2021/540} {Learning class-transductive intent representations for zero-shot intent detection}.
\newblock In \emph{IJCAI}, pages 3922--3928.

\bibitem[{Siddique et~al.(2021)Siddique, Jamour, Xu, and Hristidis}]{DBLP:conf/sigir/SiddiqueJXH21}
A.~B. Siddique, Fuad~T. Jamour, Luxun Xu, and Vagelis Hristidis. 2021.
\newblock \href {https://doi.org/10.1145/3404835.3462985} {Generalized zero-shot intent detection via commonsense knowledge}.
\newblock In \emph{SIGIR}, pages 1925--1929.

\bibitem[{Song et~al.(2023)Song, He, Wang, Dong, Mou, Wang, Xian, Cai, and Xu}]{supnid_llm_2}
Xiaoshuai Song, Keqing He, Pei Wang, Guanting Dong, Yutao Mou, Jingang Wang, Yunsen Xian, Xunliang Cai, and Weiran Xu. 2023.
\newblock \href {https://aclanthology.org/2023.emnlp-main.636} {Large language models meet open-world intent discovery and recognition: An evaluation of chatgpt}.
\newblock In \emph{EMNLP}, pages 10291--10304.

\bibitem[{Sung et~al.(2023)Sung, Gung, Mansimov, Pappas, Shu, Romeo, Zhang, and Castelli}]{emnlp/SungGM0SRZC23}
Mujeen Sung, James Gung, Elman Mansimov, Nikolaos Pappas, Raphael Shu, Salvatore Romeo, Yi~Zhang, and Vittorio Castelli. 2023.
\newblock \href {https://aclanthology.org/2023.emnlp-main.646} {Pre-training intent-aware encoders for zero- and few-shot intent classification}.
\newblock In \emph{Proceedings of EMNLP}, Singapore.

\bibitem[{Touvron et~al.(2023)Touvron, Lavril, Izacard, Martinet, Lachaux, Lacroix, Rozi{\`e}re, Goyal, Hambro, Azhar et~al.}]{touvron2023llama}
Hugo Touvron, Thibaut Lavril, Gautier Izacard, Xavier Martinet, Marie-Anne Lachaux, Timoth{\'e}e Lacroix, Baptiste Rozi{\`e}re, Naman Goyal, Eric Hambro, Faisal Azhar, et~al. 2023.
\newblock Llama: Open and efficient foundation language models.
\newblock \emph{arXiv preprint arXiv:2302.13971}.

\bibitem[{T{\"{u}}r et~al.(2011)T{\"{u}}r, Hakkani{-}T{\"{u}}r, Hillard, and Celikyilmaz}]{unsup_nvd_1}
G{\"{o}}khan T{\"{u}}r, Dilek Hakkani{-}T{\"{u}}r, Dustin Hillard, and Asli Celikyilmaz. 2011.
\newblock \href {https://doi.org/10.21437/Interspeech.2011-432} {Towards unsupervised spoken language understanding: Exploiting query click logs for slot filling}.
\newblock In \emph{INTERSPEECH}, pages 1293--1296.

\bibitem[{Vaze et~al.(2022)Vaze, Han, Vedaldi, and Zisserman}]{supnid_slm_7.6}
Sagar Vaze, Kai Han, Andrea Vedaldi, and Andrew Zisserman. 2022.
\newblock \href {https://doi.org/10.1109/CVPR52688.2022.00734} {Generalized category discovery}.
\newblock In \emph{CVPR}, pages 7482--7491.

\bibitem[{Viswanathan et~al.(2023)Viswanathan, Gashteovski, Lawrence, Wu, and Neubig}]{supnid_hybrid_2}
Vijay Viswanathan, Kiril Gashteovski, Carolin Lawrence, Tongshuang Wu, and Graham Neubig. 2023.
\newblock \href {http://arxiv.org/abs/2307.00524} {Large language models enable few-shot clustering}.

\bibitem[{Wang et~al.(2021{\natexlab{a}})Wang, Wei, Radfar, Zhang, and Chung}]{intent_dec_2}
Jixuan Wang, Kai Wei, Martin Radfar, Weiwei Zhang, and Clement Chung. 2021{\natexlab{a}}.
\newblock Encoding syntactic knowledge in transformer encoder for intent detection and slot filling.
\newblock In \emph{AAAI}, pages 13943--13951.

\bibitem[{Wang et~al.(2021{\natexlab{b}})Wang, Li, Liu, He, Yan, and Xu}]{PCLC}
Liwen Wang, Xuefeng Li, Jiachi Liu, Keqing He, Yuanmeng Yan, and Weiran Xu. 2021{\natexlab{b}}.
\newblock \href {https://doi.org/10.18653/v1/2021.emnlp-main.746} {Bridge to target domain by prototypical contrastive learning and label confusion: Re-explore zero-shot learning for slot filling}.
\newblock In \emph{EMNLP}, pages 9474--9480.

\bibitem[{Wang et~al.(2024)Wang, He, Wang, Song, Mou, Wang, Xian, Cai, and Xu}]{supnid_llm_1}
Pei Wang, Keqing He, Yejie Wang, Xiaoshuai Song, Yutao Mou, Jingang Wang, Yunsen Xian, Xunliang Cai, and Weiran Xu. 2024.
\newblock \href {https://doi.org/10.48550/arXiv.2402.17256} {Beyond the known: Investigating llms performance on out-of-domain intent detection}.
\newblock \emph{CoRR}.

\bibitem[{Wang et~al.(2019)Wang, Zhang, Li, Shi, and Jiang}]{TFWIN}
Xueying Wang, Haiqiao Zhang, Qi~Li, Yiyu Shi, and Meng Jiang. 2019.
\newblock \href {https://doi.org/10.1145/3308558.3313435} {A novel unsupervised approach for precise temporal slot filling from incomplete and noisy temporal contexts}.
\newblock In \emph{WWW}, pages 3328--3334.

\bibitem[{Wen et~al.(2017)Wen, Vandyke, Mrksic, Gasic, Rojas{-}Barahona, Su, Ultes, and Young}]{camrest_dataset}
Tsung{-}Hsien Wen, David Vandyke, Nikola Mrksic, Milica Gasic, Lina~Maria Rojas{-}Barahona, Pei{-}Hao Su, Stefan Ultes, and Steve~J. Young. 2017.
\newblock \href {https://doi.org/10.18653/v1/e17-1042} {A network-based end-to-end trainable task-oriented dialogue system}.
\newblock In \emph{EACL}, pages 438--449.

\bibitem[{Wu et~al.(2021)Wu, Su, and Juang}]{emnlp/WuSJ21}
Ting{-}Wei Wu, Ruolin Su, and Biing{-}Hwang Juang. 2021.
\newblock \href {https://doi.org/10.18653/v1/2021.emnlp-main.399} {A label-aware {BERT} attention network for zero-shot multi-intent detection in spoken language understanding}.
\newblock In \emph{EMNLP}, pages 4884--4896.

\bibitem[{Wu et~al.(2024)Wu, Dai, Zheng, and Liao}]{wu2024active}
Yuxia Wu, Tianhao Dai, Zhedong Zheng, and Lizi Liao. 2024.
\newblock Active discovering new slots for task-oriented conversation.
\newblock \emph{IEEE/ACM Transactions on Audio, Speech, and Language Processing}.

\bibitem[{Wu et~al.(2022{\natexlab{a}})Wu, Liao, Qian, and Chua}]{wu2022semi}
Yuxia Wu, Lizi Liao, Xueming Qian, and Tat-Seng Chua. 2022{\natexlab{a}}.
\newblock Semi-supervised new slot discovery with incremental clustering.
\newblock In \emph{Findings of the Association for Computational Linguistics: EMNLP 2022}, pages 6207--6218.

\bibitem[{Wu et~al.(2022{\natexlab{b}})Wu, Liao, Zhang, Lei, Zhao, Qian, and Chua}]{wu2022state}
Yuxia Wu, Lizi Liao, Gangyi Zhang, Wenqiang Lei, Guoshuai Zhao, Xueming Qian, and Tat-Seng Chua. 2022{\natexlab{b}}.
\newblock State graph reasoning for multimodal conversational recommendation.
\newblock \emph{IEEE Transactions on Multimedia}, 25:3113--3124.

\bibitem[{Xia et~al.(2018)Xia, Zhang, Yan, Chang, and Yu}]{emnlp/XiaZYCY18}
Congying Xia, Chenwei Zhang, Xiaohui Yan, Yi~Chang, and Philip~S. Yu. 2018.
\newblock \href {https://doi.org/10.18653/v1/d18-1348} {Zero-shot user intent detection via capsule neural networks}.
\newblock In \emph{EMNLP}, pages 3090--3099.

\bibitem[{Xie et~al.(2016)Xie, Girshick, and Farhadi}]{unsupnid_nn_1}
Junyuan Xie, Ross~B. Girshick, and Ali Farhadi. 2016.
\newblock \href {http://proceedings.mlr.press/v48/xieb16.html} {Unsupervised deep embedding for clustering analysis}.
\newblock In \emph{ICML}, pages 478--487.

\bibitem[{Xu et~al.(2015)Xu, Wang, Tian, Xu, Zhao, Wang, and Hao}]{stackoverflow_dataset}
Jiaming Xu, Peng Wang, Guanhua Tian, Bo~Xu, Jun Zhao, Fangyuan Wang, and Hongwei Hao. 2015.
\newblock \href {https://aclanthology.org/W15-1509} {Short text clustering via convolutional neural networks}.
\newblock In \emph{VS@HLT-NAACL}, pages 62--69.

\bibitem[{Xu et~al.(2017)Xu, Song, Zou, and Hong}]{community-based-SF}
Zengzhuang Xu, Rui Song, Bowei Zou, and Yu~Hong. 2017.
\newblock \href {https://doi.org/10.1007/978-3-319-73618-1\_54} {Unsupervised slot filler refinement via entity community construction}.
\newblock In \emph{NLPCC}, pages 642--651.

\bibitem[{Yan et~al.(2020)Yan, Fan, Li, Liu, Zhang, Wu, and Lam}]{acl/YanFLLZWL20}
Guangfeng Yan, Lu~Fan, Qimai Li, Han Liu, Xiaotong Zhang, Xiao{-}Ming Wu, and Albert Y.~S. Lam. 2020.
\newblock \href {https://doi.org/10.18653/v1/2020.acl-main.99} {Unknown intent detection using gaussian mixture model with an application to zero-shot intent classification}.
\newblock In \emph{ACL}, pages 1050--1060.

\bibitem[{Yang et~al.(2017)Yang, Fu, Sidiropoulos, and Hong}]{unsupnid_nn_2}
Bo~Yang, Xiao Fu, Nicholas~D. Sidiropoulos, and Mingyi Hong. 2017.
\newblock Towards k-means-friendly spaces: Simultaneous deep learning and clustering.
\newblock In \emph{ICML}, pages 3861--3870.

\bibitem[{Yu and Ji(2016)}]{unsup_nvd_2}
Dian Yu and Heng Ji. 2016.
\newblock \href {https://aclanthology.org/P16-1005} {Unsupervised person slot filling based on graph mining}.
\newblock In \emph{ACL}, pages 44--53.

\bibitem[{Yu et~al.(2022)Yu, Wang, Cao, Shafran, Shafey, and Soltau}]{yu2022unsupervised}
Dian Yu, Mingqiu Wang, Yuan Cao, Izhak Shafran, Laurent Shafey, and Hagen Soltau. 2022.
\newblock Unsupervised slot schema induction for task-oriented dialog.
\newblock In \emph{Proceedings of the 2022 Conference of the North American Chapter of the Association for Computational Linguistics: Human Language Technologies}, pages 1174--1193.

\bibitem[{Zeng et~al.(2021)Zeng, Ma, Yang, Gou, and Shen}]{joint_prooe_1}
Zengfeng Zeng, Dan Ma, Haiqin Yang, Zhen Gou, and Jianping Shen. 2021.
\newblock \href {https://doi.org/10.1145/3442381.3450026} {Automatic intent-slot induction for dialogue systems}.
\newblock In \emph{WWW}, pages 2578--2589.

\bibitem[{Zhang et~al.(2021{\natexlab{a}})Zhang, Nan, Wei, Li, Zhu, McKeown, Nallapati, Arnold, and Xiang}]{unsupnid_nn_5}
Dejiao Zhang, Feng Nan, Xiaokai Wei, Shang-Wen Li, Henghui Zhu, Kathleen McKeown, Ramesh Nallapati, Andrew~O. Arnold, and Bing Xiang. 2021{\natexlab{a}}.
\newblock \href {https://aclanthology.org/2021.naacl-main.427} {Supporting clustering with contrastive learning}.
\newblock In \emph{NAACL-HLT}, pages 5419--5430.

\bibitem[{Zhang et~al.(2021{\natexlab{b}})Zhang, Li, Xu, Zhang, Zhao, and Gao}]{textoir}
Hanlei Zhang, Xiaoteng Li, Hua Xu, Panpan Zhang, Kang Zhao, and Kai Gao. 2021{\natexlab{b}}.
\newblock \href {https://aclanthology.org/2021.acl-demo.20} {{TEXTOIR}: An integrated and visualized platform for text open intent recognition}.
\newblock In \emph{ACL (demo)}, pages 167--174.

\bibitem[{Zhang et~al.(2021{\natexlab{c}})Zhang, Xu, Lin, and Lyu}]{supnid_slm_6}
Hanlei Zhang, Hua Xu, Ting{-}En Lin, and Rui Lyu. 2021{\natexlab{c}}.
\newblock \href {https://doi.org/10.1609/aaai.v35i16.17689} {Discovering new intents with deep aligned clustering}.
\newblock In \emph{AAAI}, pages 14365--14373.

\bibitem[{Zhang et~al.(2023{\natexlab{a}})Zhang, Xu, Wang, Long, and Gao}]{supnid_slm_6.5}
Hanlei Zhang, Huanlin Xu, Xin Wang, Fei Long, and Kai Gao. 2023{\natexlab{a}}.
\newblock \href {https://api.semanticscholar.org/CorpusID:258179198} {A clustering framework for unsupervised and semi-supervised new intent discovery}.
\newblock \emph{TKDE}.

\bibitem[{Zhang and Zhang(2023)}]{HiCL}
Junwen Zhang and Yin Zhang. 2023.
\newblock \href {https://doi.org/10.18653/v1/2023.findings-emnlp.965} {Hierarchicalcontrast: {A} coarse-to-fine contrastive learning framework for cross-domain zero-shot slot filling}.
\newblock In \emph{Findings of EMNLP}, pages 14483--14503.

\bibitem[{Zhang et~al.(2024)Zhang, Yang, Bai, Yan, Li, Yan, and Li}]{supnid_slm_12}
Shun Zhang, Jian Yang, Jiaqi Bai, Chaoran Yan, Tongliang Li, Zhao Yan, and Zhoujun Li. 2024.
\newblock \href {https://aclanthology.org/2024.lrec-main.1067} {New intent discovery with attracting and dispersing prototype}.
\newblock In \emph{LREC-COLING}, pages 12193--12206.

\bibitem[{Zhang et~al.(2023{\natexlab{b}})Zhang, Wang, and Shang}]{supnid_hybrid_1}
Yuwei Zhang, Zihan Wang, and Jingbo Shang. 2023{\natexlab{b}}.
\newblock \href {https://aclanthology.org/2023.emnlp-main.858} {Clusterllm: Large language models as a guide for text clustering}.
\newblock In \emph{EMNLP}, pages 13903--13920.

\bibitem[{Zhang et~al.(2022)Zhang, Zhang, Zhan, Wu, and Lam}]{supnid_slm_8}
Yuwei Zhang, Haode Zhang, Li{-}Ming Zhan, Xiao{-}Ming Wu, and Albert Y.~S. Lam. 2022.
\newblock \href {https://doi.org/10.18653/v1/2022.acl-long.21} {New intent discovery with pre-training and contrastive learning}.
\newblock In \emph{ACL}, pages 256--269.

\bibitem[{Zhang et~al.(2019)Zhang, Liao, Huang, Zhu, and Chua}]{zhang2019neural}
Zheng Zhang, Lizi Liao, Minlie Huang, Xiaoyan Zhu, and Tat-Seng Chua. 2019.
\newblock Neural multimodal belief tracker with adaptive attention for dialogue systems.
\newblock In \emph{The world wide web conference}, pages 2401--2412.

\bibitem[{Zhao and Feng(2018)}]{nvd_based_2}
Lin Zhao and Zhe Feng. 2018.
\newblock \href {https://aclanthology.org/P18-2068/} {Improving slot filling in spoken language understanding with joint pointer and attention}.
\newblock In \emph{ACL}, pages 426--431.

\bibitem[{Zhou et~al.(2023)Zhou, Quan, and Qiu}]{supnid_slm_9}
Yunhua Zhou, Guofeng Quan, and Xipeng Qiu. 2023.
\newblock \href {https://doi.org/10.18653/v1/2023.acl-long.209} {A probabilistic framework for discovering new intents}.
\newblock In \emph{ACL}, pages 3771--3784.

\bibitem[{Zhu et~al.(2020)Zhu, Wang, Li, Wu, He, and Zhou}]{multimodal_1}
Tiangang Zhu, Yue Wang, Haoran Li, Youzheng Wu, Xiaodong He, and Bowen Zhou. 2020.
\newblock \href {https://doi.org/10.18653/v1/2020.emnlp-main.166} {Multimodal joint attribute prediction and value extraction for e-commerce product}.
\newblock In \emph{EMNLP}, pages 2129--2139.

\end{thebibliography}
\newpage
\appendix

\clearpage
\section{Appendix}
\label{sec:appendix}

\subsection{Data Resources}
\label{app:data_resources}

\paragraph{New Intent Discovery Datasets.} We show three widely used datasets for NID. Specifically, \textbf{BANGING77} \citep{banking_dataset} is a fine-grained intent discovery dataset sourced from banking domain dialogues. It contains over 13K user utterances distributed across 77 unique intents. \textbf{CLINC150} \citep{clinc_dataset}, on the other hand, is a multi-domain dataset featuring 150 distinct intents and 22,500 utterances across 10 different domains. \textbf{StackOverflow} \citep{stackoverflow_dataset}, a dataset curated from Kaggle.com, includes 20,000 technical questions categorized into 20 distinct areas.

\paragraph{New Slot-Value Discovery Datasets.} For the NSVD task, we introduce seven prominent datasets spanning various domains. The \textbf{CamRest} dataset, provided by \citet{camrest_dataset}, delves into the restaurant domain, boasting over 2,700 utterances across 4 slots, offering valuable insights into task-oriented dialogues. Similarly, the \textbf{Cambridge SLU} dataset by \citet{canbridge_slu_dataset} also explores the restaurant sector, featuring more than 10,500 utterances across 5 slots. Additionally, the MultiWOZ dataset spans multiple domains, with its subsets, \textbf{WOZ-attr} \citep{multiwoz/EricGPSAGKGKH20} and \textbf{WOZ-hotel} \citep{multiwoz/EricGPSAGKGKH20}, exploring the attraction and hotel domains with over 7,500 and 14,000 utterances, respectively. Despite encompassing intents, the limited intent quantity in these datasets restricts their suitability for the NID task. Conversely, the \textbf{ATIS} dataset \citep{atis_dataset} expands into the flight domain with nearly 5,000 utterances and 120 slots. The \textbf{SNIPS} dataset \citep{Snips_dataset} provides a valuable resource for spoken language understanding across seven domains, boasting 72 slots and around 2,000 utterances per domain.  The \textbf{SGD} \citep{SGD/RastogiZSGK20} contains dialogues from 16 domains with a total of 46 intents and 214 slots. Notably, ATIS, SNIPS, and SGD are replete with a variety of intents, thus making them apt for comprehensive studies in both NID and NSVD tasks.

\begin{table*}[tp]
\centering
\resizebox{1.0\linewidth}{!}{
\begin{tabular}{c|lllllllll}
\toprule
\multirow{3}{*}[3pt]{Methods} & \multicolumn{3}{c}{\textbf{BANKING77}} & \multicolumn{3}{c}{\textbf{CLINC150}} & \multicolumn{3}{c}{\textbf{StackOverflow}} \\ \cline{2-10}
&         &         &         &          &         &         &           &           &           \\[-6pt]
& \textbf{ACC} & ARI & NMI & \textbf{ACC} & ARI & NMI & \textbf{ACC} & ARI & NMI       \\
\midrule
\multicolumn{10}{c}{\textit{Statistical Methods}} \\ % 添加空白行
\midrule
K-Means \citep{unsupnid_stat_6} & 29.55 & 12.18 & 54.57 & 45.06 & 26.86 & 70.89 & 13.55 & 1.46 & 8.24 \\
AG   \citep{unsupnid_stat_7}    & 31.58 & 13.31 & 57.07 & 44.03 & 27.70 & 73.07 & 14.66 & 2.12 & 10.62 \\
\midrule
\multicolumn{10}{c}{\textit{NN-based Methods}} \\ % 添加空白行
\midrule
DEC   \citep{unsupnid_nn_1}     & 41.29 & 27.21 & 67.78 & 46.89 & 27.46 & 74.83 & 13.09 & 3.76 & 10.88 \\
DCN   \citep{unsupnid_nn_2}     & 41.99 & 26.81 & 67.54 & 49.29 & 31.15 & 75.66 & 34.26 & 15.45 & 31.09 \\
DAC   \citep{unsupnid_nn_3}     & 27.41 & 14.24 & 47.35 & 55.94 & 40.49 & 78.40 & 16.30 & 2.76 & 14.71 \\
DeepCluster \citep{unsupnid_nn_4}& 20.69 & 8.95  & 41.77 & 35.70 & 19.11 & 65.58 & -& -    & -     \\
SCCL   \citep{unsupnid_nn_5}    & 40.54 & 26.98 & 63.89 & 50.44 & 38.14 & 79.35 & 68.15 & 34.81 & 69.11 \\
USNID \citep{supnid_slm_6.5}  & 54.83 & 43.33 & 75.30 & 75.87 & 68.54 & 91.00 & 69.28 & 52.25 & 72.00   \\
IDAS  \citep{unsupnid_nn_6}   & 67.43 & 57.56 & 82.84 & 85.48 & 79.02 & 93.82 & 83.82 & 72.20 & 81.26 \\
\bottomrule
\end{tabular}
} 
% \vspace{-0.1cm}
\caption{The main unsupervised NID results on three benchmarks.}
\label{tab:nid_unsupervised}
% \vspace{-0.2cm}
\end{table*}

\subsection{Evaluation Protocols}
\label{app:eval_protocols}

\begin{table*}[!htp]
\centering
\resizebox{1.0\linewidth}{!}{
\begin{tabular}{c|ccccc}
\toprule
\multirow{2}{*}[3pt]{Methods} & \textbf{CamRest} & \textbf{Cambridge SLU} & \textbf{WOZ-hotel}  & \textbf{WOZ-attr}   & \textbf{ATIS}\\ %\cline{1-6}
 
% \midrule
% \multicolumn{6}{c}{\textit{Unsupervised Methods}} \\ % 添加空白行

\midrule
 DistFrame-Sem\citep{chen2014leveraging} & 53.5    &59.0      &38.2      &37.5       &61.6     \\
Merge-Select\citep{hudevcek2021discovering} &55.2   &66.4      & 38.8      & 38.3  &64.8   \\
% \midrule

% \multicolumn{6}{c}{\textit{weakly-supervised Methods}} \\ % 添加空白行
% \midrule
%  \textit{Merge-Select\citep{hudevcek2021discovering}}  & 0.665   & 0.692  & 0.548  &0.439   & 0.710   \\
\bottomrule
\end{tabular}
}
% \vspace{-0.1cm}
\caption{The main results of unsupervised NSVD methods on five benchmarks. Here we provide the Span-F1 score.}
\label{tab:nsd_unsupervised}
% \vspace{-0.2cm}
\end{table*}

\paragraph{NID Metrics.} The NID task involves accurately assigning utterances into their corresponding intent groups from potentially many possibilities. Accordingly, the performance of NID models is typically assessed using three standard metrics: ACC, ARI, and NMI \citep{supnid_slm_6, supnid_slm_8}, which evaluate how effectively the model identifies and groups intents, ensuring that the clustering reflects true user intentions rather than random associations. As previously mentioned, ACC assesses NID performance by calculating the proportion of correctly predicted outputs to total predictions, aligned with ground-truth labels. Notably, the ACC in this context is derived following an alignment process using the Hungarian algorithm. The definition of ACC is as follows:
\begin{equation}
    ACC = \frac{\sum_{i=1}^{N} \mathbbm{1}_{y_i=map(\hat{y_i})}}{N},
\end{equation}
where $\{\hat{y_i}, y_i\}$ denote the predicted and true labels, respectively. $map(\cdot)$ is the Hungarian algorithm-based mapping function.

Different from ACC, ARI measures the concordance of the predicted and actual clusters through an assessment of pairwise accuracy within clusters, which is computed as:
\begin{equation}
\resizebox{0.9\linewidth}{!}{$ARI = \frac{\sum_{i, j} \tbinom{n_{i,j}}{2} - [\sum_{i} \tbinom{u_{i}}{2}\sum_{j} \tbinom{v_{j}}{2}]/\tbinom{N}{2}}{\frac{1}{2} [\sum_{i} \tbinom{u_{i}}{2} + \sum_{j} \tbinom{v_{j}}{2}] - [\sum_{i} \tbinom{u_{i}}{2}\sum_{j} \tbinom{v_{j}}{2}]/\tbinom{N}{2}}$},
\end{equation}
where $n_{i, j}$ denotes the number of sample pairs both in $i^{th}$ predicted and $j^{th}$ ground-truth cluster. $u_i = \sum_{j}n_{i,j}$, and  $v_j = \sum_{i}n_{i,j}$ represent the sum of sample pairs in the same predicted and true clusters, respectively. $N$ is the number of all samples.

Regarding the NMI, it aims to gauge the level of agreement between the predicted and ground-truth clusters by quantifying the normalized mutual information between them. It can be calculated as follows:
\begin{equation}
    NMI(\hat{\boldsymbol{y}}, \boldsymbol{y}) = \frac{2 \cdot I(\hat{\boldsymbol{y}}, \boldsymbol{y})}{H(\hat{\boldsymbol{y}}) + H(\boldsymbol{y})},
\end{equation}
where $\{\hat{\boldsymbol{y}}, \boldsymbol{y}\}$ denote the predicted labels and the ground-truth labels respectively. $I(\cdot)$ signifies mutual information. $H(\cdot)$ is the entropy function.

\paragraph{NSVD Metrics.} For the NSVD task, the challenge lies in accurately identifying relevant slots and values within utterances and precisely delineating their boundaries. Metrics such as Precision, Recall, and Span-F1 are essential for assessing the performance of NSVD models. These metrics ensure the accuracy and completeness of information extraction, focusing on specific elements within utterances. Considering a set of actual slot values \(M_1, M_2, \ldots, M_n\), where \(n\) is the number of slots, and a corresponding set of predicted values \(\varepsilon_1, \varepsilon_2, \ldots, \varepsilon_n\), precision \(P_i\) and recall \(R_i\) are calculated for each slot type \(i\) as follows:

\begin{equation}
P_i = \frac{|M_i \cap \varepsilon_i|}{|\varepsilon_i|},
\end{equation}

\begin{equation}
R_i = \frac{|M_i \cap \varepsilon_i|}{|M_i|}.
\end{equation}

The overall weighted precision $P$ and recall $R$ are computed as follows:

\begin{table*}[!htp]
\centering
\resizebox{1.0\linewidth}{!}{
\begin{tabular}{lccccccc}
\toprule
& \multicolumn{5}{c}{Sequence tagging-based models} & \multicolumn{1}{c}{MRC-based models} & \multicolumn{1}{c}{Prompting-based models} \\
\cmidrule(lr){2-6} \cmidrule(lr){7-7} \cmidrule(lr){8-8}
Domain & CT & RZT & Coach & CZSL & PCLC & RCSF & GZPL \\
\midrule
AddToPlaylist         & 38.82 & 42.77 & 50.90 & 53.89 & 59.24 & 68.70 & 61.64 \\
BookRestaurant        & 27.54 & 30.68 & 34.01 & 34.06 & 41.36 & 63.49 & 62.93 \\
GetWeather            & 46.45 & 50.28 & 50.47 & 52.04 & 54.21 & 65.36 & 64.97 \\
PlayMusic             & 32.86 & 33.12 & 32.01 & 34.59 & 34.95 & 53.51 & 66.42 \\
RateBook              & 14.54 & 16.43 & 22.06 & 31.53 & 29.31 & 36.51 & 47.53 \\
SearchCreativeWork    & 39.79 & 44.45 & 46.65 & 50.61 & 53.51 & 69.22 & 72.88 \\
SearchScreeningEvent  & 13.83 & 12.25 & 25.63 & 30.05 & 27.17 & 33.54 & 51.42 \\
\midrule
Average F1            & 30.55 & 32.85 & 37.39 & 40.99 & 42.82 & 55.76 & 61.07 \\
\bottomrule
\end{tabular}
}
\vspace{-0.1cm}
\caption{The main results of Partially Supervised NSVD methods the SNIPS dataset.}
\label{tab:nvd_semi_supervised}
\vspace{-0.2cm}
\end{table*}

\begin{equation}
P = \frac{\sum_{i=1}^{n} |\varepsilon_i| P_i}{\sum_{j=1}^{n} |\varepsilon_j|},
\end{equation}

\begin{equation}
R = \frac{\sum_{i=1}^{n} |M_i| R_i}{\sum_{j=1}^{n} |M_j|}.
\end{equation}

The F1 score is then computed as the harmonic mean of the overall weighted precision and recall, thus accounting for both the precision and recall in a balanced manner:

\begin{equation}
F1 = \frac{2PR}{P + R}.
\end{equation}

In the context of slot value spans, this metric is specifically referred to as Span-F1.

\paragraph{Other Metrics.} While NID and NSVD metrics offer valuable insights into OnExp model performance, their uniform application across all test data can obscure distinctions between utterances containing known versus novel ontological items. To address this, metrics such as \textbf{Known ACC}, \textbf{Novel ACC}, and the \textbf{H-score} are indispensable, as they effectively differentiate model performance on known and novel items, providing a more granular assessment of model capabilities \citep{supnid_slm_11}. Specifically, \textbf{Known ACC} and \textbf{Novel ACC} are specialized forms of \textbf{ACC}, computed separately for known and novel ontological items. The \textbf{H-score} is calculated as the harmonic mean of \textbf{Known ACC} and \textbf{Novel ACC} as follows:
\begin{equation}
\scalebox{1.1}{ % Change 1.5 to whatever scale factor you need
$\text{H-score} = \frac{2}{ 1/\text{Know ACC} + 1/\text{Novel ACC}}$.
}\end{equation}

\section{Leaderboard}
\label{app:leaderboard}

\noindent \textbf{NID Leaderboard.} Table \ref{tab:nid_unsupervised} presents the unsupervised NID results on three benchmarks. Notably, although USNID is categorized into the semi-supervised NID methods, it can adapt to an unsupervised setting. Hence, we have included USNID results in the unsupervised context for a comprehensive evaluation.

\vspace{0.3em}
\noindent \textbf{NSVD Leaderboard.} Table \ref{tab:nsd_unsupervised} and Table \ref{tab:nvd_semi_supervised} present the main performance of unsupervised NSVD methods and partially supervised NSVD methods.

We adopted results reported in the published literature \citep{supnid_slm_6, supnid_slm_6.5, supnid_slm_9, supnid_slm_12, supnid_slm_13, wu2022semi, wu2024active}.

\end{document}